\documentclass[letterpaper,10pt,conference]{ieeeconf}
\IEEEoverridecommandlockouts
\usepackage{amsmath,amssymb,booktabs,graphicx,xcolor,caption,import,microtype,needspace,placeins}
\usepackage[normalem]{ulem}
\usepackage{multirow}
\usepackage[hidelinks]{hyperref}
\usepackage[subtle,tracking=normal,wordspacing=normal]{savetrees}
\definecolor{reviewblue}{HTML}{175BBB}
\definecolor{reviewpurple}{HTML}{783C9D}
\definecolor{cutgreen}{HTML}{267440}
\newif\ifshowchanges
\showchangesfalse
\ifdefined\ForceReviewMarkup\showchangestrue\fi
\ifshowchanges\colorlet{red}{reviewblue}\fi
\ifshowchanges\else\colorlet{reviewpurple}{black}\fi

\renewenvironment{itemize}{\begin{list}{$\bullet$}{\setlength{\leftmargin}{10pt}\setlength{\itemsep}{1pt}\setlength{\parsep}{0pt}\setlength{\topsep}{3pt}}}{\end{list}}
\renewcommand{\paragraph}[1]{\par\Needspace{4\baselineskip}\smallskip\noindent\textit{#1}\par\nobreak}
\expandafter\def\csname released-base-Tracking\endcsname{5.47}
\expandafter\def\csname released-base-Violation\endcsname{4.13}
\expandafter\def\csname released-base-TV\endcsname{0.205}
\expandafter\def\csname released-base-Falls\endcsname{13}
\expandafter\def\csname released-base-Depth\endcsname{5.12}
\expandafter\def\csname released-base-Incidence\endcsname{46.61}
\expandafter\def\csname blind-filter-off-zero-Tracking\endcsname{5.26}
\expandafter\def\csname blind-filter-off-zero-Violation\endcsname{5.11}
\expandafter\def\csname blind-filter-off-zero-TV\endcsname{0.224}
\expandafter\def\csname blind-filter-off-zero-Falls\endcsname{21}
\expandafter\def\csname blind-filter-off-zero-Depth\endcsname{7.00}
\expandafter\def\csname blind-filter-off-zero-Incidence\endcsname{54.96}
\expandafter\def\csname blind-filter-on-zero-Tracking\endcsname{6.04}
\expandafter\def\csname blind-filter-on-zero-Violation\endcsname{7.94}
\expandafter\def\csname blind-filter-on-zero-TV\endcsname{0.327}
\expandafter\def\csname blind-filter-on-zero-Falls\endcsname{9}
\expandafter\def\csname blind-filter-on-zero-Depth\endcsname{7.68}
\expandafter\def\csname blind-filter-on-zero-Incidence\endcsname{69.60}
\expandafter\def\csname full-zero-Tracking\endcsname{6.03}
\expandafter\def\csname full-zero-Violation\endcsname{2.20}
\expandafter\def\csname full-zero-TV\endcsname{0.361}
\expandafter\def\csname full-zero-Falls\endcsname{9}
\expandafter\def\csname full-zero-Depth\endcsname{5.41}
\expandafter\def\csname full-zero-Incidence\endcsname{35.23}
\expandafter\def\csname full-delta-cr-Tracking\endcsname{6.07}
\expandafter\def\csname full-delta-cr-Violation\endcsname{0.86}
\expandafter\def\csname full-delta-cr-TV\endcsname{0.271}
\expandafter\def\csname full-delta-cr-Falls\endcsname{8}
\expandafter\def\csname full-delta-cr-Depth\endcsname{2.58}
\expandafter\def\csname full-delta-cr-Incidence\endcsname{17.80}
\expandafter\def\csname late-crs-Tracking\endcsname{6.16}
\expandafter\def\csname late-crs-Violation\endcsname{0.70}
\expandafter\def\csname late-crs-TV\endcsname{0.280}
\expandafter\def\csname late-crs-Falls\endcsname{6}
\expandafter\def\csname late-crs-Depth\endcsname{2.20}
\expandafter\def\csname late-crs-Incidence\endcsname{15.45}
\expandafter\def\csname early-cr-Tracking\endcsname{6.07}
\expandafter\def\csname early-cr-Violation\endcsname{0.64}
\expandafter\def\csname early-cr-TV\endcsname{0.283}
\expandafter\def\csname early-cr-Falls\endcsname{5}
\expandafter\def\csname early-cr-Depth\endcsname{1.98}
\expandafter\def\csname early-cr-Incidence\endcsname{14.23}
\expandafter\def\csname early-crs-Tracking\endcsname{6.31}
\expandafter\def\csname early-crs-Violation\endcsname{0.60}
\expandafter\def\csname early-crs-TV\endcsname{0.288}
\expandafter\def\csname early-crs-Falls\endcsname{7}
\expandafter\def\csname early-crs-Depth\endcsname{1.91}
\expandafter\def\csname early-crs-Incidence\endcsname{13.94}
\expandafter\def\csname bound-only-cr-Tracking\endcsname{6.41}
\expandafter\def\csname bound-only-cr-Violation\endcsname{13.02}
\expandafter\def\csname bound-only-cr-TV\endcsname{0.254}
\expandafter\def\csname bound-only-cr-Falls\endcsname{7}
\expandafter\def\csname bound-only-cr-Depth\endcsname{7.07}
\expandafter\def\csname bound-only-cr-Incidence\endcsname{76.15}
\expandafter\def\csname direction-only-cr-Tracking\endcsname{6.42}
\expandafter\def\csname direction-only-cr-Violation\endcsname{10.63}
\expandafter\def\csname direction-only-cr-TV\endcsname{0.263}
\expandafter\def\csname direction-only-cr-Falls\endcsname{9}
\expandafter\def\csname direction-only-cr-Depth\endcsname{7.61}
\expandafter\def\csname direction-only-cr-Incidence\endcsname{67.72}
\expandafter\def\csname h-lfh-cr-Tracking\endcsname{6.15}
\expandafter\def\csname h-lfh-cr-Violation\endcsname{8.40}
\expandafter\def\csname h-lfh-cr-TV\endcsname{0.229}
\expandafter\def\csname h-lfh-cr-Falls\endcsname{12}
\expandafter\def\csname h-lfh-cr-Depth\endcsname{5.43}
\expandafter\def\csname h-lfh-cr-Incidence\endcsname{66.99}
\expandafter\def\csname action-bound-cr-Tracking\endcsname{6.34}
\expandafter\def\csname action-bound-cr-Violation\endcsname{5.04}
\expandafter\def\csname action-bound-cr-TV\endcsname{0.262}
\expandafter\def\csname action-bound-cr-Falls\endcsname{9}
\expandafter\def\csname action-bound-cr-Depth\endcsname{5.39}
\expandafter\def\csname action-bound-cr-Incidence\endcsname{53.71}
\expandafter\def\csname delta-only-cr-Tracking\endcsname{5.76}
\expandafter\def\csname delta-only-cr-Violation\endcsname{9.79}
\expandafter\def\csname delta-only-cr-TV\endcsname{0.221}
\expandafter\def\csname delta-only-cr-Falls\endcsname{9}
\expandafter\def\csname delta-only-cr-Depth\endcsname{5.15}
\expandafter\def\csname delta-only-cr-Incidence\endcsname{71.07}
\expandafter\def\csname historical-full-cr-Tracking\endcsname{6.23}
\expandafter\def\csname historical-full-cr-Violation\endcsname{1.05}
\expandafter\def\csname historical-full-cr-TV\endcsname{0.276}
\expandafter\def\csname historical-full-cr-Falls\endcsname{9}
\expandafter\def\csname historical-full-cr-Depth\endcsname{2.96}
\expandafter\def\csname historical-full-cr-Incidence\endcsname{20.24}
\expandafter\def\csname top8-history-cr-Tracking\endcsname{6.04}
\expandafter\def\csname top8-history-cr-Violation\endcsname{1.61}
\expandafter\def\csname top8-history-cr-TV\endcsname{0.229}
\expandafter\def\csname top8-history-cr-Falls\endcsname{10}
\expandafter\def\csname top8-history-cr-Depth\endcsname{3.34}
\expandafter\def\csname top8-history-cr-Incidence\endcsname{25.88}
\expandafter\def\csname masked-max-no-delta-cr-Tracking\endcsname{6.05}
\expandafter\def\csname masked-max-no-delta-cr-Violation\endcsname{1.38}
\expandafter\def\csname masked-max-no-delta-cr-TV\endcsname{0.191}
\expandafter\def\csname masked-max-no-delta-cr-Falls\endcsname{12}
\expandafter\def\csname masked-max-no-delta-cr-Depth\endcsname{2.92}
\expandafter\def\csname masked-max-no-delta-cr-Incidence\endcsname{20.97}
\expandafter\def\csname softmax-delta-cr-Tracking\endcsname{5.84}
\expandafter\def\csname softmax-delta-cr-Violation\endcsname{2.13}
\expandafter\def\csname softmax-delta-cr-TV\endcsname{0.197}
\expandafter\def\csname softmax-delta-cr-Falls\endcsname{8}
\expandafter\def\csname softmax-delta-cr-Depth\endcsname{3.37}
\expandafter\def\csname softmax-delta-cr-Incidence\endcsname{30.69}
\expandafter\def\csname attention-late-delta-cr-Tracking\endcsname{5.83}
\expandafter\def\csname attention-late-delta-cr-Violation\endcsname{2.56}
\expandafter\def\csname attention-late-delta-cr-TV\endcsname{0.204}
\expandafter\def\csname attention-late-delta-cr-Falls\endcsname{8}
\expandafter\def\csname attention-late-delta-cr-Depth\endcsname{3.50}
\expandafter\def\csname attention-late-delta-cr-Incidence\endcsname{34.01}
\expandafter\def\csname sphere-oracle-cr-Tracking\endcsname{5.79}
\expandafter\def\csname sphere-oracle-cr-Violation\endcsname{1.40}
\expandafter\def\csname sphere-oracle-cr-TV\endcsname{0.199}
\expandafter\def\csname sphere-oracle-cr-Falls\endcsname{9}
\expandafter\def\csname sphere-oracle-cr-Depth\endcsname{3.02}
\expandafter\def\csname sphere-oracle-cr-Incidence\endcsname{22.73}
\expandafter\def\csname current-only-cr-Tracking\endcsname{5.84}
\expandafter\def\csname current-only-cr-Violation\endcsname{1.21}
\expandafter\def\csname current-only-cr-TV\endcsname{0.271}
\expandafter\def\csname current-only-cr-Falls\endcsname{8}
\expandafter\def\csname current-only-cr-Depth\endcsname{2.96}
\expandafter\def\csname current-only-cr-Incidence\endcsname{21.92}
\expandafter\def\csname full-cr-Tracking\endcsname{6.19}
\expandafter\def\csname full-cr-Violation\endcsname{1.07}
\expandafter\def\csname full-cr-TV\endcsname{0.276}
\expandafter\def\csname full-cr-Falls\endcsname{9}
\expandafter\def\csname full-cr-Depth\endcsname{2.91}
\expandafter\def\csname full-cr-Incidence\endcsname{20.34}
\expandafter\def\csname full-c-Tracking\endcsname{6.00}
\expandafter\def\csname full-c-Violation\endcsname{1.39}
\expandafter\def\csname full-c-TV\endcsname{0.266}
\expandafter\def\csname full-c-Falls\endcsname{9}
\expandafter\def\csname full-c-Depth\endcsname{3.34}
\expandafter\def\csname full-c-Incidence\endcsname{24.07}
\expandafter\def\csname full-crj-Tracking\endcsname{6.29}
\expandafter\def\csname full-crj-Violation\endcsname{0.90}
\expandafter\def\csname full-crj-TV\endcsname{0.283}
\expandafter\def\csname full-crj-Falls\endcsname{12}
\expandafter\def\csname full-crj-Depth\endcsname{2.57}
\expandafter\def\csname full-crj-Incidence\endcsname{18.43}
\expandafter\def\csname full-crh-Tracking\endcsname{6.31}
\expandafter\def\csname full-crh-Violation\endcsname{0.97}
\expandafter\def\csname full-crh-TV\endcsname{0.276}
\expandafter\def\csname full-crh-Falls\endcsname{8}
\expandafter\def\csname full-crh-Depth\endcsname{2.57}
\expandafter\def\csname full-crh-Incidence\endcsname{19.63}
\expandafter\def\csname full-crs-Tracking\endcsname{6.28}
\expandafter\def\csname full-crs-Violation\endcsname{0.95}
\expandafter\def\csname full-crs-TV\endcsname{0.275}
\expandafter\def\csname full-crs-Falls\endcsname{11}
\expandafter\def\csname full-crs-Depth\endcsname{2.86}
\expandafter\def\csname full-crs-Incidence\endcsname{18.92}
\expandafter\def\csname released-base-Off\endcsname{4.75}
\expandafter\def\csname early-crs-Off\endcsname{5.69}
\expandafter\def\csname twist2-hardware-full-ViolationDp2\endcsname{8.13}
\expandafter\def\csname twist2-hardware-full-ViolationInt\endcsname{8}
\expandafter\def\csname twist2-hardware-cpt-ViolationDp2\endcsname{46.85}
\expandafter\def\csname twist2-hardware-cpt-ViolationInt\endcsname{47}
\expandafter\def\csname sonic-hardware-full-ViolationDp2\endcsname{5.24}
\expandafter\def\csname sonic-hardware-full-ViolationInt\endcsname{5}
\expandafter\def\csname sonic-hardware-cpt-ViolationDp2\endcsname{7.64}
\expandafter\def\csname sonic-hardware-cpt-ViolationInt\endcsname{8}
\expandafter\def\csname full-cr-vs-action-bound-cr-Violation-RelativeDropDp2\endcsname{78.77}
\expandafter\def\csname full-cr-vs-action-bound-cr-Violation-RelativeDropInt\endcsname{79}
\expandafter\def\csname full-cr-vs-action-bound-cr-Violation-RelativeIncreaseDp2\endcsname{-78.77}
\expandafter\def\csname full-cr-vs-action-bound-cr-Violation-RelativeIncreaseInt\endcsname{-79}
\expandafter\def\csname full-cr-vs-action-bound-cr-Violation-DeltaPctDp2\endcsname{-78.77}
\expandafter\def\csname full-cr-vs-action-bound-cr-Violation-DeltaPctInt\endcsname{-79}
\expandafter\def\csname full-cr-vs-action-bound-cr-Violation-DeltaPPDp2\endcsname{-3.97}
\expandafter\def\csname full-cr-vs-action-bound-cr-Violation-DeltaPPInt\endcsname{-4}
\expandafter\def\csname early-crs-vs-released-base-Violation-RelativeDropDp2\endcsname{85.47}
\expandafter\def\csname early-crs-vs-released-base-Violation-RelativeDropInt\endcsname{85}
\expandafter\def\csname early-crs-vs-released-base-Violation-RelativeIncreaseDp2\endcsname{-85.47}
\expandafter\def\csname early-crs-vs-released-base-Violation-RelativeIncreaseInt\endcsname{-85}
\expandafter\def\csname early-crs-vs-released-base-Violation-DeltaPctDp2\endcsname{-85.47}
\expandafter\def\csname early-crs-vs-released-base-Violation-DeltaPctInt\endcsname{-85}
\expandafter\def\csname early-crs-vs-released-base-Violation-DeltaPPDp2\endcsname{-3.53}
\expandafter\def\csname early-crs-vs-released-base-Violation-DeltaPPInt\endcsname{-4}
\expandafter\def\csname early-crs-vs-released-base-Tracking-RelativeDropDp2\endcsname{-15.36}
\expandafter\def\csname early-crs-vs-released-base-Tracking-RelativeDropInt\endcsname{-15}
\expandafter\def\csname early-crs-vs-released-base-Tracking-RelativeIncreaseDp2\endcsname{15.36}
\expandafter\def\csname early-crs-vs-released-base-Tracking-RelativeIncreaseInt\endcsname{15}
\expandafter\def\csname twist2-scenes-full-vs-twist2-scenes-filter-only-Violation-RelativeDropDp2\endcsname{91.14}
\expandafter\def\csname twist2-scenes-full-vs-twist2-scenes-filter-only-Violation-RelativeDropInt\endcsname{91}
\expandafter\def\csname twist2-scenes-full-vs-twist2-scenes-filter-only-Violation-RelativeIncreaseDp2\endcsname{-91.14}
\expandafter\def\csname twist2-scenes-full-vs-twist2-scenes-filter-only-Violation-RelativeIncreaseInt\endcsname{-91}
\expandafter\def\csname twist2-scenes-full-vs-twist2-scenes-filter-only-Violation-DeltaPctDp2\endcsname{-91.14}
\expandafter\def\csname twist2-scenes-full-vs-twist2-scenes-filter-only-Violation-DeltaPctInt\endcsname{-91}
\expandafter\def\csname twist2-scenes-full-vs-twist2-scenes-filter-only-Violation-DeltaPPDp2\endcsname{-17.60}
\expandafter\def\csname twist2-scenes-full-vs-twist2-scenes-filter-only-Violation-DeltaPPInt\endcsname{-18}
\expandafter\def\csname twist2-scenes-full-vs-twist2-scenes-cpt-Violation-RelativeDropDp2\endcsname{92.80}
\expandafter\def\csname twist2-scenes-full-vs-twist2-scenes-cpt-Violation-RelativeDropInt\endcsname{93}
\expandafter\def\csname twist2-scenes-full-vs-twist2-scenes-cpt-Violation-RelativeIncreaseDp2\endcsname{-92.80}
\expandafter\def\csname twist2-scenes-full-vs-twist2-scenes-cpt-Violation-RelativeIncreaseInt\endcsname{-93}
\expandafter\def\csname twist2-scenes-full-vs-twist2-scenes-cpt-Violation-DeltaPctDp2\endcsname{-92.80}
\expandafter\def\csname twist2-scenes-full-vs-twist2-scenes-cpt-Violation-DeltaPctInt\endcsname{-93}
\expandafter\def\csname twist2-scenes-full-vs-twist2-scenes-cpt-Violation-DeltaPPDp2\endcsname{-22.04}
\expandafter\def\csname twist2-scenes-full-vs-twist2-scenes-cpt-Violation-DeltaPPInt\endcsname{-22}
\expandafter\def\csname twist2-scenes-observations-vs-twist2-scenes-cpt-Violation-RelativeDropDp2\endcsname{78.91}
\expandafter\def\csname twist2-scenes-observations-vs-twist2-scenes-cpt-Violation-RelativeDropInt\endcsname{79}
\expandafter\def\csname twist2-scenes-observations-vs-twist2-scenes-cpt-Violation-RelativeIncreaseDp2\endcsname{-78.91}
\expandafter\def\csname twist2-scenes-observations-vs-twist2-scenes-cpt-Violation-RelativeIncreaseInt\endcsname{-79}
\expandafter\def\csname twist2-scenes-observations-vs-twist2-scenes-cpt-Violation-DeltaPctDp2\endcsname{-78.91}
\expandafter\def\csname twist2-scenes-observations-vs-twist2-scenes-cpt-Violation-DeltaPctInt\endcsname{-79}
\expandafter\def\csname twist2-scenes-observations-vs-twist2-scenes-cpt-Violation-DeltaPPDp2\endcsname{-18.74}
\expandafter\def\csname twist2-scenes-observations-vs-twist2-scenes-cpt-Violation-DeltaPPInt\endcsname{-19}
\expandafter\def\csname twist2-scenes-rewards-vs-twist2-scenes-cpt-Violation-RelativeDropDp2\endcsname{34.82}
\expandafter\def\csname twist2-scenes-rewards-vs-twist2-scenes-cpt-Violation-RelativeDropInt\endcsname{35}
\expandafter\def\csname twist2-scenes-rewards-vs-twist2-scenes-cpt-Violation-RelativeIncreaseDp2\endcsname{-34.82}
\expandafter\def\csname twist2-scenes-rewards-vs-twist2-scenes-cpt-Violation-RelativeIncreaseInt\endcsname{-35}
\expandafter\def\csname twist2-scenes-rewards-vs-twist2-scenes-cpt-Violation-DeltaPctDp2\endcsname{-34.82}
\expandafter\def\csname twist2-scenes-rewards-vs-twist2-scenes-cpt-Violation-DeltaPctInt\endcsname{-35}
\expandafter\def\csname twist2-scenes-rewards-vs-twist2-scenes-cpt-Violation-DeltaPPDp2\endcsname{-8.27}
\expandafter\def\csname twist2-scenes-rewards-vs-twist2-scenes-cpt-Violation-DeltaPPInt\endcsname{-8}
\expandafter\def\csname twist2-scenes-filter-only-vs-twist2-scenes-cpt-Violation-RelativeDropDp2\endcsname{18.69}
\expandafter\def\csname twist2-scenes-filter-only-vs-twist2-scenes-cpt-Violation-RelativeDropInt\endcsname{19}
\expandafter\def\csname twist2-scenes-filter-only-vs-twist2-scenes-cpt-Violation-RelativeIncreaseDp2\endcsname{-18.69}
\expandafter\def\csname twist2-scenes-filter-only-vs-twist2-scenes-cpt-Violation-RelativeIncreaseInt\endcsname{-19}
\expandafter\def\csname twist2-scenes-filter-only-vs-twist2-scenes-cpt-Violation-DeltaPctDp2\endcsname{-18.69}
\expandafter\def\csname twist2-scenes-filter-only-vs-twist2-scenes-cpt-Violation-DeltaPctInt\endcsname{-19}
\expandafter\def\csname twist2-scenes-filter-only-vs-twist2-scenes-cpt-Violation-DeltaPPDp2\endcsname{-4.44}
\expandafter\def\csname twist2-scenes-filter-only-vs-twist2-scenes-cpt-Violation-DeltaPPInt\endcsname{-4}
\expandafter\def\csname twist2-hardware-full-vs-twist2-hardware-cpt-Violation-RelativeDropDp2\endcsname{82.65}
\expandafter\def\csname twist2-hardware-full-vs-twist2-hardware-cpt-Violation-RelativeDropInt\endcsname{83}
\expandafter\def\csname twist2-hardware-full-vs-twist2-hardware-cpt-Violation-RelativeIncreaseDp2\endcsname{-82.65}
\expandafter\def\csname twist2-hardware-full-vs-twist2-hardware-cpt-Violation-RelativeIncreaseInt\endcsname{-83}
\expandafter\def\csname twist2-hardware-full-vs-twist2-hardware-cpt-Violation-DeltaPctDp2\endcsname{-82.65}
\expandafter\def\csname twist2-hardware-full-vs-twist2-hardware-cpt-Violation-DeltaPctInt\endcsname{-83}
\expandafter\def\csname twist2-hardware-full-vs-twist2-hardware-cpt-Violation-DeltaPPDp2\endcsname{-38.72}
\expandafter\def\csname twist2-hardware-full-vs-twist2-hardware-cpt-Violation-DeltaPPInt\endcsname{-39}
\expandafter\def\csname sonic-scenes-full-vs-sonic-scenes-filter-only-Violation-RelativeDropDp2\endcsname{20.63}
\expandafter\def\csname sonic-scenes-full-vs-sonic-scenes-filter-only-Violation-RelativeDropInt\endcsname{21}
\expandafter\def\csname sonic-scenes-full-vs-sonic-scenes-filter-only-Violation-RelativeIncreaseDp2\endcsname{-20.63}
\expandafter\def\csname sonic-scenes-full-vs-sonic-scenes-filter-only-Violation-RelativeIncreaseInt\endcsname{-21}
\expandafter\def\csname sonic-scenes-full-vs-sonic-scenes-filter-only-Violation-DeltaPctDp2\endcsname{-20.63}
\expandafter\def\csname sonic-scenes-full-vs-sonic-scenes-filter-only-Violation-DeltaPctInt\endcsname{-21}
\expandafter\def\csname sonic-scenes-full-vs-sonic-scenes-filter-only-Violation-DeltaPPDp2\endcsname{-2.93}
\expandafter\def\csname sonic-scenes-full-vs-sonic-scenes-filter-only-Violation-DeltaPPInt\endcsname{-3}
\expandafter\def\csname sonic-scenes-full-vs-sonic-scenes-cpt-Violation-RelativeDropDp2\endcsname{44.89}
\expandafter\def\csname sonic-scenes-full-vs-sonic-scenes-cpt-Violation-RelativeDropInt\endcsname{45}
\expandafter\def\csname sonic-scenes-full-vs-sonic-scenes-cpt-Violation-RelativeIncreaseDp2\endcsname{-44.89}
\expandafter\def\csname sonic-scenes-full-vs-sonic-scenes-cpt-Violation-RelativeIncreaseInt\endcsname{-45}
\expandafter\def\csname sonic-scenes-full-vs-sonic-scenes-cpt-Violation-DeltaPctDp2\endcsname{-44.89}
\expandafter\def\csname sonic-scenes-full-vs-sonic-scenes-cpt-Violation-DeltaPctInt\endcsname{-45}
\expandafter\def\csname sonic-scenes-full-vs-sonic-scenes-cpt-Violation-DeltaPPDp2\endcsname{-9.18}
\expandafter\def\csname sonic-scenes-full-vs-sonic-scenes-cpt-Violation-DeltaPPInt\endcsname{-9}
\expandafter\def\csname sonic-scenes-observations-vs-sonic-scenes-cpt-Violation-RelativeDropDp2\endcsname{39.36}
\expandafter\def\csname sonic-scenes-observations-vs-sonic-scenes-cpt-Violation-RelativeDropInt\endcsname{39}
\expandafter\def\csname sonic-scenes-observations-vs-sonic-scenes-cpt-Violation-RelativeIncreaseDp2\endcsname{-39.36}
\expandafter\def\csname sonic-scenes-observations-vs-sonic-scenes-cpt-Violation-RelativeIncreaseInt\endcsname{-39}
\expandafter\def\csname sonic-scenes-observations-vs-sonic-scenes-cpt-Violation-DeltaPctDp2\endcsname{-39.36}
\expandafter\def\csname sonic-scenes-observations-vs-sonic-scenes-cpt-Violation-DeltaPctInt\endcsname{-39}
\expandafter\def\csname sonic-scenes-observations-vs-sonic-scenes-cpt-Violation-DeltaPPDp2\endcsname{-8.05}
\expandafter\def\csname sonic-scenes-observations-vs-sonic-scenes-cpt-Violation-DeltaPPInt\endcsname{-8}
\expandafter\def\csname sonic-scenes-rewards-vs-sonic-scenes-cpt-Violation-RelativeDropDp2\endcsname{34.57}
\expandafter\def\csname sonic-scenes-rewards-vs-sonic-scenes-cpt-Violation-RelativeDropInt\endcsname{35}
\expandafter\def\csname sonic-scenes-rewards-vs-sonic-scenes-cpt-Violation-RelativeIncreaseDp2\endcsname{-34.57}
\expandafter\def\csname sonic-scenes-rewards-vs-sonic-scenes-cpt-Violation-RelativeIncreaseInt\endcsname{-35}
\expandafter\def\csname sonic-scenes-rewards-vs-sonic-scenes-cpt-Violation-DeltaPctDp2\endcsname{-34.57}
\expandafter\def\csname sonic-scenes-rewards-vs-sonic-scenes-cpt-Violation-DeltaPctInt\endcsname{-35}
\expandafter\def\csname sonic-scenes-rewards-vs-sonic-scenes-cpt-Violation-DeltaPPDp2\endcsname{-7.07}
\expandafter\def\csname sonic-scenes-rewards-vs-sonic-scenes-cpt-Violation-DeltaPPInt\endcsname{-7}
\expandafter\def\csname sonic-scenes-filter-only-vs-sonic-scenes-cpt-Violation-RelativeDropDp2\endcsname{30.56}
\expandafter\def\csname sonic-scenes-filter-only-vs-sonic-scenes-cpt-Violation-RelativeDropInt\endcsname{31}
\expandafter\def\csname sonic-scenes-filter-only-vs-sonic-scenes-cpt-Violation-RelativeIncreaseDp2\endcsname{-30.56}
\expandafter\def\csname sonic-scenes-filter-only-vs-sonic-scenes-cpt-Violation-RelativeIncreaseInt\endcsname{-31}
\expandafter\def\csname sonic-scenes-filter-only-vs-sonic-scenes-cpt-Violation-DeltaPctDp2\endcsname{-30.56}
\expandafter\def\csname sonic-scenes-filter-only-vs-sonic-scenes-cpt-Violation-DeltaPctInt\endcsname{-31}
\expandafter\def\csname sonic-scenes-filter-only-vs-sonic-scenes-cpt-Violation-DeltaPPDp2\endcsname{-6.25}
\expandafter\def\csname sonic-scenes-filter-only-vs-sonic-scenes-cpt-Violation-DeltaPPInt\endcsname{-6}
\expandafter\def\csname sonic-hardware-full-vs-sonic-hardware-cpt-Violation-RelativeDropDp2\endcsname{31.41}
\expandafter\def\csname sonic-hardware-full-vs-sonic-hardware-cpt-Violation-RelativeDropInt\endcsname{31}
\expandafter\def\csname sonic-hardware-full-vs-sonic-hardware-cpt-Violation-RelativeIncreaseDp2\endcsname{-31.41}
\expandafter\def\csname sonic-hardware-full-vs-sonic-hardware-cpt-Violation-RelativeIncreaseInt\endcsname{-31}
\expandafter\def\csname sonic-hardware-full-vs-sonic-hardware-cpt-Violation-DeltaPctDp2\endcsname{-31.41}
\expandafter\def\csname sonic-hardware-full-vs-sonic-hardware-cpt-Violation-DeltaPctInt\endcsname{-31}
\expandafter\def\csname sonic-hardware-full-vs-sonic-hardware-cpt-Violation-DeltaPPDp2\endcsname{-2.40}
\expandafter\def\csname sonic-hardware-full-vs-sonic-hardware-cpt-Violation-DeltaPPInt\endcsname{-2}
\expandafter\def\csname released-base-vs-blind-filter-off-zero-Violation-RelativeDropDp2\endcsname{19.18}
\expandafter\def\csname released-base-vs-blind-filter-off-zero-Violation-RelativeDropInt\endcsname{19}
\expandafter\def\csname released-base-vs-blind-filter-off-zero-Violation-RelativeIncreaseDp2\endcsname{-19.18}
\expandafter\def\csname released-base-vs-blind-filter-off-zero-Violation-RelativeIncreaseInt\endcsname{-19}
\expandafter\def\csname released-base-vs-blind-filter-off-zero-Violation-DeltaPctDp2\endcsname{-19.18}
\expandafter\def\csname released-base-vs-blind-filter-off-zero-Violation-DeltaPctInt\endcsname{-19}
\expandafter\def\csname released-base-vs-blind-filter-off-zero-Violation-DeltaPPDp2\endcsname{-0.98}
\expandafter\def\csname released-base-vs-blind-filter-off-zero-Violation-DeltaPPInt\endcsname{-1}
\expandafter\def\csname blind-filter-on-zero-vs-blind-filter-off-zero-Violation-RelativeDropDp2\endcsname{-55.38}
\expandafter\def\csname blind-filter-on-zero-vs-blind-filter-off-zero-Violation-RelativeDropInt\endcsname{-55}
\expandafter\def\csname blind-filter-on-zero-vs-blind-filter-off-zero-Violation-RelativeIncreaseDp2\endcsname{55.38}
\expandafter\def\csname blind-filter-on-zero-vs-blind-filter-off-zero-Violation-RelativeIncreaseInt\endcsname{55}
\expandafter\def\csname blind-filter-on-zero-vs-blind-filter-off-zero-Violation-DeltaPctDp2\endcsname{55.38}
\expandafter\def\csname blind-filter-on-zero-vs-blind-filter-off-zero-Violation-DeltaPctInt\endcsname{55}
\expandafter\def\csname blind-filter-on-zero-vs-blind-filter-off-zero-Violation-DeltaPPDp2\endcsname{2.83}
\expandafter\def\csname blind-filter-on-zero-vs-blind-filter-off-zero-Violation-DeltaPPInt\endcsname{3}
\expandafter\def\csname full-zero-vs-blind-filter-off-zero-Violation-RelativeDropDp2\endcsname{56.95}
\expandafter\def\csname full-zero-vs-blind-filter-off-zero-Violation-RelativeDropInt\endcsname{57}
\expandafter\def\csname full-zero-vs-blind-filter-off-zero-Violation-RelativeIncreaseDp2\endcsname{-56.95}
\expandafter\def\csname full-zero-vs-blind-filter-off-zero-Violation-RelativeIncreaseInt\endcsname{-57}
\expandafter\def\csname full-zero-vs-blind-filter-off-zero-Violation-DeltaPctDp2\endcsname{-56.95}
\expandafter\def\csname full-zero-vs-blind-filter-off-zero-Violation-DeltaPctInt\endcsname{-57}
\expandafter\def\csname full-zero-vs-blind-filter-off-zero-Violation-DeltaPPDp2\endcsname{-2.91}
\expandafter\def\csname full-zero-vs-blind-filter-off-zero-Violation-DeltaPPInt\endcsname{-3}
\expandafter\def\csname full-delta-cr-vs-blind-filter-off-zero-Violation-RelativeDropDp2\endcsname{83.17}
\expandafter\def\csname full-delta-cr-vs-blind-filter-off-zero-Violation-RelativeDropInt\endcsname{83}
\expandafter\def\csname full-delta-cr-vs-blind-filter-off-zero-Violation-RelativeIncreaseDp2\endcsname{-83.17}
\expandafter\def\csname full-delta-cr-vs-blind-filter-off-zero-Violation-RelativeIncreaseInt\endcsname{-83}
\expandafter\def\csname full-delta-cr-vs-blind-filter-off-zero-Violation-DeltaPctDp2\endcsname{-83.17}
\expandafter\def\csname full-delta-cr-vs-blind-filter-off-zero-Violation-DeltaPctInt\endcsname{-83}
\expandafter\def\csname full-delta-cr-vs-blind-filter-off-zero-Violation-DeltaPPDp2\endcsname{-4.25}
\expandafter\def\csname full-delta-cr-vs-blind-filter-off-zero-Violation-DeltaPPInt\endcsname{-4}
\expandafter\def\csname late-crs-vs-blind-filter-off-zero-Violation-RelativeDropDp2\endcsname{86.30}
\expandafter\def\csname late-crs-vs-blind-filter-off-zero-Violation-RelativeDropInt\endcsname{86}
\expandafter\def\csname late-crs-vs-blind-filter-off-zero-Violation-RelativeIncreaseDp2\endcsname{-86.30}
\expandafter\def\csname late-crs-vs-blind-filter-off-zero-Violation-RelativeIncreaseInt\endcsname{-86}
\expandafter\def\csname late-crs-vs-blind-filter-off-zero-Violation-DeltaPctDp2\endcsname{-86.30}
\expandafter\def\csname late-crs-vs-blind-filter-off-zero-Violation-DeltaPctInt\endcsname{-86}
\expandafter\def\csname late-crs-vs-blind-filter-off-zero-Violation-DeltaPPDp2\endcsname{-4.41}
\expandafter\def\csname late-crs-vs-blind-filter-off-zero-Violation-DeltaPPInt\endcsname{-4}
\expandafter\def\csname early-cr-vs-blind-filter-off-zero-Violation-RelativeDropDp2\endcsname{87.48}
\expandafter\def\csname early-cr-vs-blind-filter-off-zero-Violation-RelativeDropInt\endcsname{87}
\expandafter\def\csname early-cr-vs-blind-filter-off-zero-Violation-RelativeIncreaseDp2\endcsname{-87.48}
\expandafter\def\csname early-cr-vs-blind-filter-off-zero-Violation-RelativeIncreaseInt\endcsname{-87}
\expandafter\def\csname early-cr-vs-blind-filter-off-zero-Violation-DeltaPctDp2\endcsname{-87.48}
\expandafter\def\csname early-cr-vs-blind-filter-off-zero-Violation-DeltaPctInt\endcsname{-87}
\expandafter\def\csname early-cr-vs-blind-filter-off-zero-Violation-DeltaPPDp2\endcsname{-4.47}
\expandafter\def\csname early-cr-vs-blind-filter-off-zero-Violation-DeltaPPInt\endcsname{-4}
\expandafter\def\csname early-crs-vs-blind-filter-off-zero-Violation-RelativeDropDp2\endcsname{88.26}
\expandafter\def\csname early-crs-vs-blind-filter-off-zero-Violation-RelativeDropInt\endcsname{88}
\expandafter\def\csname early-crs-vs-blind-filter-off-zero-Violation-RelativeIncreaseDp2\endcsname{-88.26}
\expandafter\def\csname early-crs-vs-blind-filter-off-zero-Violation-RelativeIncreaseInt\endcsname{-88}
\expandafter\def\csname early-crs-vs-blind-filter-off-zero-Violation-DeltaPctDp2\endcsname{-88.26}
\expandafter\def\csname early-crs-vs-blind-filter-off-zero-Violation-DeltaPctInt\endcsname{-88}
\expandafter\def\csname early-crs-vs-blind-filter-off-zero-Violation-DeltaPPDp2\endcsname{-4.51}
\expandafter\def\csname early-crs-vs-blind-filter-off-zero-Violation-DeltaPPInt\endcsname{-5}
\expandafter\def\csname bound-only-cr-vs-blind-filter-off-zero-Violation-RelativeDropDp2\endcsname{-154.79}
\expandafter\def\csname bound-only-cr-vs-blind-filter-off-zero-Violation-RelativeDropInt\endcsname{-155}
\expandafter\def\csname bound-only-cr-vs-blind-filter-off-zero-Violation-RelativeIncreaseDp2\endcsname{154.79}
\expandafter\def\csname bound-only-cr-vs-blind-filter-off-zero-Violation-RelativeIncreaseInt\endcsname{155}
\expandafter\def\csname bound-only-cr-vs-blind-filter-off-zero-Violation-DeltaPctDp2\endcsname{154.79}
\expandafter\def\csname bound-only-cr-vs-blind-filter-off-zero-Violation-DeltaPctInt\endcsname{155}
\expandafter\def\csname bound-only-cr-vs-blind-filter-off-zero-Violation-DeltaPPDp2\endcsname{7.91}
\expandafter\def\csname bound-only-cr-vs-blind-filter-off-zero-Violation-DeltaPPInt\endcsname{8}
\expandafter\def\csname direction-only-cr-vs-blind-filter-off-zero-Violation-RelativeDropDp2\endcsname{-108.02}
\expandafter\def\csname direction-only-cr-vs-blind-filter-off-zero-Violation-RelativeDropInt\endcsname{-108}
\expandafter\def\csname direction-only-cr-vs-blind-filter-off-zero-Violation-RelativeIncreaseDp2\endcsname{108.02}
\expandafter\def\csname direction-only-cr-vs-blind-filter-off-zero-Violation-RelativeIncreaseInt\endcsname{108}
\expandafter\def\csname direction-only-cr-vs-blind-filter-off-zero-Violation-DeltaPctDp2\endcsname{108.02}
\expandafter\def\csname direction-only-cr-vs-blind-filter-off-zero-Violation-DeltaPctInt\endcsname{108}
\expandafter\def\csname direction-only-cr-vs-blind-filter-off-zero-Violation-DeltaPPDp2\endcsname{5.52}
\expandafter\def\csname direction-only-cr-vs-blind-filter-off-zero-Violation-DeltaPPInt\endcsname{6}
\expandafter\def\csname h-lfh-cr-vs-blind-filter-off-zero-Violation-RelativeDropDp2\endcsname{-64.38}
\expandafter\def\csname h-lfh-cr-vs-blind-filter-off-zero-Violation-RelativeDropInt\endcsname{-64}
\expandafter\def\csname h-lfh-cr-vs-blind-filter-off-zero-Violation-RelativeIncreaseDp2\endcsname{64.38}
\expandafter\def\csname h-lfh-cr-vs-blind-filter-off-zero-Violation-RelativeIncreaseInt\endcsname{64}
\expandafter\def\csname h-lfh-cr-vs-blind-filter-off-zero-Violation-DeltaPctDp2\endcsname{64.38}
\expandafter\def\csname h-lfh-cr-vs-blind-filter-off-zero-Violation-DeltaPctInt\endcsname{64}
\expandafter\def\csname h-lfh-cr-vs-blind-filter-off-zero-Violation-DeltaPPDp2\endcsname{3.29}
\expandafter\def\csname h-lfh-cr-vs-blind-filter-off-zero-Violation-DeltaPPInt\endcsname{3}
\expandafter\def\csname action-bound-cr-vs-blind-filter-off-zero-Violation-RelativeDropDp2\endcsname{1.37}
\expandafter\def\csname action-bound-cr-vs-blind-filter-off-zero-Violation-RelativeDropInt\endcsname{1}
\expandafter\def\csname action-bound-cr-vs-blind-filter-off-zero-Violation-RelativeIncreaseDp2\endcsname{-1.37}
\expandafter\def\csname action-bound-cr-vs-blind-filter-off-zero-Violation-RelativeIncreaseInt\endcsname{-1}
\expandafter\def\csname action-bound-cr-vs-blind-filter-off-zero-Violation-DeltaPctDp2\endcsname{-1.37}
\expandafter\def\csname action-bound-cr-vs-blind-filter-off-zero-Violation-DeltaPctInt\endcsname{-1}
\expandafter\def\csname action-bound-cr-vs-blind-filter-off-zero-Violation-DeltaPPDp2\endcsname{-0.07}
\expandafter\def\csname action-bound-cr-vs-blind-filter-off-zero-Violation-DeltaPPInt\endcsname{-0}
\expandafter\def\csname delta-only-cr-vs-blind-filter-off-zero-Violation-RelativeDropDp2\endcsname{-91.59}
\expandafter\def\csname delta-only-cr-vs-blind-filter-off-zero-Violation-RelativeDropInt\endcsname{-92}
\expandafter\def\csname delta-only-cr-vs-blind-filter-off-zero-Violation-RelativeIncreaseDp2\endcsname{91.59}
\expandafter\def\csname delta-only-cr-vs-blind-filter-off-zero-Violation-RelativeIncreaseInt\endcsname{92}
\expandafter\def\csname delta-only-cr-vs-blind-filter-off-zero-Violation-DeltaPctDp2\endcsname{91.59}
\expandafter\def\csname delta-only-cr-vs-blind-filter-off-zero-Violation-DeltaPctInt\endcsname{92}
\expandafter\def\csname delta-only-cr-vs-blind-filter-off-zero-Violation-DeltaPPDp2\endcsname{4.68}
\expandafter\def\csname delta-only-cr-vs-blind-filter-off-zero-Violation-DeltaPPInt\endcsname{5}
\expandafter\def\csname historical-full-cr-vs-blind-filter-off-zero-Violation-RelativeDropDp2\endcsname{79.45}
\expandafter\def\csname historical-full-cr-vs-blind-filter-off-zero-Violation-RelativeDropInt\endcsname{79}
\expandafter\def\csname historical-full-cr-vs-blind-filter-off-zero-Violation-RelativeIncreaseDp2\endcsname{-79.45}
\expandafter\def\csname historical-full-cr-vs-blind-filter-off-zero-Violation-RelativeIncreaseInt\endcsname{-79}
\expandafter\def\csname historical-full-cr-vs-blind-filter-off-zero-Violation-DeltaPctDp2\endcsname{-79.45}
\expandafter\def\csname historical-full-cr-vs-blind-filter-off-zero-Violation-DeltaPctInt\endcsname{-79}
\expandafter\def\csname historical-full-cr-vs-blind-filter-off-zero-Violation-DeltaPPDp2\endcsname{-4.06}
\expandafter\def\csname historical-full-cr-vs-blind-filter-off-zero-Violation-DeltaPPInt\endcsname{-4}
\expandafter\def\csname top8-history-cr-vs-blind-filter-off-zero-Violation-RelativeDropDp2\endcsname{68.49}
\expandafter\def\csname top8-history-cr-vs-blind-filter-off-zero-Violation-RelativeDropInt\endcsname{68}
\expandafter\def\csname top8-history-cr-vs-blind-filter-off-zero-Violation-RelativeIncreaseDp2\endcsname{-68.49}
\expandafter\def\csname top8-history-cr-vs-blind-filter-off-zero-Violation-RelativeIncreaseInt\endcsname{-68}
\expandafter\def\csname top8-history-cr-vs-blind-filter-off-zero-Violation-DeltaPctDp2\endcsname{-68.49}
\expandafter\def\csname top8-history-cr-vs-blind-filter-off-zero-Violation-DeltaPctInt\endcsname{-68}
\expandafter\def\csname top8-history-cr-vs-blind-filter-off-zero-Violation-DeltaPPDp2\endcsname{-3.50}
\expandafter\def\csname top8-history-cr-vs-blind-filter-off-zero-Violation-DeltaPPInt\endcsname{-4}
\expandafter\def\csname masked-max-no-delta-cr-vs-blind-filter-off-zero-Violation-RelativeDropDp2\endcsname{72.99}
\expandafter\def\csname masked-max-no-delta-cr-vs-blind-filter-off-zero-Violation-RelativeDropInt\endcsname{73}
\expandafter\def\csname masked-max-no-delta-cr-vs-blind-filter-off-zero-Violation-RelativeIncreaseDp2\endcsname{-72.99}
\expandafter\def\csname masked-max-no-delta-cr-vs-blind-filter-off-zero-Violation-RelativeIncreaseInt\endcsname{-73}
\expandafter\def\csname masked-max-no-delta-cr-vs-blind-filter-off-zero-Violation-DeltaPctDp2\endcsname{-72.99}
\expandafter\def\csname masked-max-no-delta-cr-vs-blind-filter-off-zero-Violation-DeltaPctInt\endcsname{-73}
\expandafter\def\csname masked-max-no-delta-cr-vs-blind-filter-off-zero-Violation-DeltaPPDp2\endcsname{-3.73}
\expandafter\def\csname masked-max-no-delta-cr-vs-blind-filter-off-zero-Violation-DeltaPPInt\endcsname{-4}
\expandafter\def\csname softmax-delta-cr-vs-blind-filter-off-zero-Violation-RelativeDropDp2\endcsname{58.32}
\expandafter\def\csname softmax-delta-cr-vs-blind-filter-off-zero-Violation-RelativeDropInt\endcsname{58}
\expandafter\def\csname softmax-delta-cr-vs-blind-filter-off-zero-Violation-RelativeIncreaseDp2\endcsname{-58.32}
\expandafter\def\csname softmax-delta-cr-vs-blind-filter-off-zero-Violation-RelativeIncreaseInt\endcsname{-58}
\expandafter\def\csname softmax-delta-cr-vs-blind-filter-off-zero-Violation-DeltaPctDp2\endcsname{-58.32}
\expandafter\def\csname softmax-delta-cr-vs-blind-filter-off-zero-Violation-DeltaPctInt\endcsname{-58}
\expandafter\def\csname softmax-delta-cr-vs-blind-filter-off-zero-Violation-DeltaPPDp2\endcsname{-2.98}
\expandafter\def\csname softmax-delta-cr-vs-blind-filter-off-zero-Violation-DeltaPPInt\endcsname{-3}
\expandafter\def\csname attention-late-delta-cr-vs-blind-filter-off-zero-Violation-RelativeDropDp2\endcsname{49.90}
\expandafter\def\csname attention-late-delta-cr-vs-blind-filter-off-zero-Violation-RelativeDropInt\endcsname{50}
\expandafter\def\csname attention-late-delta-cr-vs-blind-filter-off-zero-Violation-RelativeIncreaseDp2\endcsname{-49.90}
\expandafter\def\csname attention-late-delta-cr-vs-blind-filter-off-zero-Violation-RelativeIncreaseInt\endcsname{-50}
\expandafter\def\csname attention-late-delta-cr-vs-blind-filter-off-zero-Violation-DeltaPctDp2\endcsname{-49.90}
\expandafter\def\csname attention-late-delta-cr-vs-blind-filter-off-zero-Violation-DeltaPctInt\endcsname{-50}
\expandafter\def\csname attention-late-delta-cr-vs-blind-filter-off-zero-Violation-DeltaPPDp2\endcsname{-2.55}
\expandafter\def\csname attention-late-delta-cr-vs-blind-filter-off-zero-Violation-DeltaPPInt\endcsname{-3}
\expandafter\def\csname sphere-oracle-cr-vs-blind-filter-off-zero-Violation-RelativeDropDp2\endcsname{72.60}
\expandafter\def\csname sphere-oracle-cr-vs-blind-filter-off-zero-Violation-RelativeDropInt\endcsname{73}
\expandafter\def\csname sphere-oracle-cr-vs-blind-filter-off-zero-Violation-RelativeIncreaseDp2\endcsname{-72.60}
\expandafter\def\csname sphere-oracle-cr-vs-blind-filter-off-zero-Violation-RelativeIncreaseInt\endcsname{-73}
\expandafter\def\csname sphere-oracle-cr-vs-blind-filter-off-zero-Violation-DeltaPctDp2\endcsname{-72.60}
\expandafter\def\csname sphere-oracle-cr-vs-blind-filter-off-zero-Violation-DeltaPctInt\endcsname{-73}
\expandafter\def\csname sphere-oracle-cr-vs-blind-filter-off-zero-Violation-DeltaPPDp2\endcsname{-3.71}
\expandafter\def\csname sphere-oracle-cr-vs-blind-filter-off-zero-Violation-DeltaPPInt\endcsname{-4}
\expandafter\def\csname current-only-cr-vs-blind-filter-off-zero-Violation-RelativeDropDp2\endcsname{76.32}
\expandafter\def\csname current-only-cr-vs-blind-filter-off-zero-Violation-RelativeDropInt\endcsname{76}
\expandafter\def\csname current-only-cr-vs-blind-filter-off-zero-Violation-RelativeIncreaseDp2\endcsname{-76.32}
\expandafter\def\csname current-only-cr-vs-blind-filter-off-zero-Violation-RelativeIncreaseInt\endcsname{-76}
\expandafter\def\csname current-only-cr-vs-blind-filter-off-zero-Violation-DeltaPctDp2\endcsname{-76.32}
\expandafter\def\csname current-only-cr-vs-blind-filter-off-zero-Violation-DeltaPctInt\endcsname{-76}
\expandafter\def\csname current-only-cr-vs-blind-filter-off-zero-Violation-DeltaPPDp2\endcsname{-3.90}
\expandafter\def\csname current-only-cr-vs-blind-filter-off-zero-Violation-DeltaPPInt\endcsname{-4}
\expandafter\def\csname full-cr-vs-blind-filter-off-zero-Violation-RelativeDropDp2\endcsname{79.06}
\expandafter\def\csname full-cr-vs-blind-filter-off-zero-Violation-RelativeDropInt\endcsname{79}
\expandafter\def\csname full-cr-vs-blind-filter-off-zero-Violation-RelativeIncreaseDp2\endcsname{-79.06}
\expandafter\def\csname full-cr-vs-blind-filter-off-zero-Violation-RelativeIncreaseInt\endcsname{-79}
\expandafter\def\csname full-cr-vs-blind-filter-off-zero-Violation-DeltaPctDp2\endcsname{-79.06}
\expandafter\def\csname full-cr-vs-blind-filter-off-zero-Violation-DeltaPctInt\endcsname{-79}
\expandafter\def\csname full-cr-vs-blind-filter-off-zero-Violation-DeltaPPDp2\endcsname{-4.04}
\expandafter\def\csname full-cr-vs-blind-filter-off-zero-Violation-DeltaPPInt\endcsname{-4}
\expandafter\def\csname full-c-vs-blind-filter-off-zero-Violation-RelativeDropDp2\endcsname{72.80}
\expandafter\def\csname full-c-vs-blind-filter-off-zero-Violation-RelativeDropInt\endcsname{73}
\expandafter\def\csname full-c-vs-blind-filter-off-zero-Violation-RelativeIncreaseDp2\endcsname{-72.80}
\expandafter\def\csname full-c-vs-blind-filter-off-zero-Violation-RelativeIncreaseInt\endcsname{-73}
\expandafter\def\csname full-c-vs-blind-filter-off-zero-Violation-DeltaPctDp2\endcsname{-72.80}
\expandafter\def\csname full-c-vs-blind-filter-off-zero-Violation-DeltaPctInt\endcsname{-73}
\expandafter\def\csname full-c-vs-blind-filter-off-zero-Violation-DeltaPPDp2\endcsname{-3.72}
\expandafter\def\csname full-c-vs-blind-filter-off-zero-Violation-DeltaPPInt\endcsname{-4}
\expandafter\def\csname full-crj-vs-blind-filter-off-zero-Violation-RelativeDropDp2\endcsname{82.39}
\expandafter\def\csname full-crj-vs-blind-filter-off-zero-Violation-RelativeDropInt\endcsname{82}
\expandafter\def\csname full-crj-vs-blind-filter-off-zero-Violation-RelativeIncreaseDp2\endcsname{-82.39}
\expandafter\def\csname full-crj-vs-blind-filter-off-zero-Violation-RelativeIncreaseInt\endcsname{-82}
\expandafter\def\csname full-crj-vs-blind-filter-off-zero-Violation-DeltaPctDp2\endcsname{-82.39}
\expandafter\def\csname full-crj-vs-blind-filter-off-zero-Violation-DeltaPctInt\endcsname{-82}
\expandafter\def\csname full-crj-vs-blind-filter-off-zero-Violation-DeltaPPDp2\endcsname{-4.21}
\expandafter\def\csname full-crj-vs-blind-filter-off-zero-Violation-DeltaPPInt\endcsname{-4}
\expandafter\def\csname full-crh-vs-blind-filter-off-zero-Violation-RelativeDropDp2\endcsname{81.02}
\expandafter\def\csname full-crh-vs-blind-filter-off-zero-Violation-RelativeDropInt\endcsname{81}
\expandafter\def\csname full-crh-vs-blind-filter-off-zero-Violation-RelativeIncreaseDp2\endcsname{-81.02}
\expandafter\def\csname full-crh-vs-blind-filter-off-zero-Violation-RelativeIncreaseInt\endcsname{-81}
\expandafter\def\csname full-crh-vs-blind-filter-off-zero-Violation-DeltaPctDp2\endcsname{-81.02}
\expandafter\def\csname full-crh-vs-blind-filter-off-zero-Violation-DeltaPctInt\endcsname{-81}
\expandafter\def\csname full-crh-vs-blind-filter-off-zero-Violation-DeltaPPDp2\endcsname{-4.14}
\expandafter\def\csname full-crh-vs-blind-filter-off-zero-Violation-DeltaPPInt\endcsname{-4}
\expandafter\def\csname full-crs-vs-blind-filter-off-zero-Violation-RelativeDropDp2\endcsname{81.41}
\expandafter\def\csname full-crs-vs-blind-filter-off-zero-Violation-RelativeDropInt\endcsname{81}
\expandafter\def\csname full-crs-vs-blind-filter-off-zero-Violation-RelativeIncreaseDp2\endcsname{-81.41}
\expandafter\def\csname full-crs-vs-blind-filter-off-zero-Violation-RelativeIncreaseInt\endcsname{-81}
\expandafter\def\csname full-crs-vs-blind-filter-off-zero-Violation-DeltaPctDp2\endcsname{-81.41}
\expandafter\def\csname full-crs-vs-blind-filter-off-zero-Violation-DeltaPctInt\endcsname{-81}
\expandafter\def\csname full-crs-vs-blind-filter-off-zero-Violation-DeltaPPDp2\endcsname{-4.16}
\expandafter\def\csname full-crs-vs-blind-filter-off-zero-Violation-DeltaPPInt\endcsname{-4}

\newcommand{\val}[2]{\csname #1-#2\endcsname}

\newcommand{\sd}[1]{\ensuremath{_{\mbox{\scriptsize#1}}}}

\title{\LARGE\bf Filter-Aware Fine-Tuning
for Safe Humanoid Whole-Body Tracking}
\author{Pranit Mohnot$^{1}$, Christian Helten$^{2}$, Daniele Gammelli$^{1,3}$, Marco Pavone$^{1,4}$%
\thanks{$^{1}$Stanford University, $^{2}$Technical University of Munich, $^{3}$Italian Institute of Artificial Intelligence (AI4I), $^{4}$NVIDIA}
}

\begin{document}
\maketitle
\thispagestyle{empty}\pagestyle{empty}
\begin{abstract}
Safe whole-body motion is essential for deploying humanoid robots in unstructured environments. Modern humanoid control commonly separates reference specification from execution, with a planner, teleoperator, or motion generator providing a reference that a reinforcement-learning policy tracks through dynamically feasible whole-body control. Runtime safety filters, such as control barrier functions (CBFs), offer a promising approach for enforcing newly introduced constraints via interventions on the tracker's outputs. We show, however, that treating the tracking policy and safety filter independently induces fundamental mismatches, as filtering alters both the executed actions and the induced state distribution. We study this policy–filter interface through case studies that isolate dynamics, objective, and information mismatches, highlight their root causes, and use these insights to develop \textit{CoFiT} (\underline{Co}nstrained \underline{Fi}lter-aware \underline{T}uning), a filter-aware fine-tuning method for pretrained trackers. Across diverse constraint scenes, CoFiT reduces violation time relative to filter-only training by 91\% on TWIST2 and 21\% on SONIC, while requiring smaller safety filter corrections. On Unitree G1 hardware, CoFiT reduces violation time by 83\% for TWIST2 and completes every trial without operator intervention, whereas 50\% of baseline trials require an operator stop. Together, these results provide actionable insights into policy–filter interactions and establish design principles for integrating learned trackers with runtime safety filters.

\end{abstract}

\section{Introduction}
\label{sec:introduction}

Humanoid robots are rapidly moving toward deployment in homes, warehouses, hospitals, and other human-centered environments, where safe operation requires coordinating the full body under task- and environment-dependent constraints.
Whole-body control increasingly relies on learning-based motion tracking, in which a motion source, such as a teleoperator, recorded demonstration, or kinematic motion generator, provides a reference trajectory that a reinforcement-learning (RL) tracking policy translates into joint commands consistent with the robot’s dynamics~\cite{ze2025twist2,luo2026sonic,he2024omnih2o}.
While this control architecture enables expressive whole-body motion, ensuring its safety in real-world environments remains challenging.
A reference trajectory specifies the intended motion but does not, in general, encode every constraint of the environment in which that motion will be executed.
Obstacles, workspace limits, and other safety requirements may be absent from the motion source, emerge only at runtime, or change after the tracking policy has been trained. 

Two broad strategies have emerged for enforcing safety in whole-body control.
The first internalizes safety within the learned tracking policy, for example by training with penalties on constraint violations or barrier-based rewards that encourage avoidance of unsafe states~\cite{yang2026pacman}.
This approach tightly couples safety to training, as accommodating a new constraint generally requires retraining, and the resulting policy does not by itself provide a runtime guarantee.
The second strategy places a safety filter downstream of the tracking policy to correct its actions at runtime~\cite{ames2017control,wabersich2021a}. 
This modular approach leaves the nominal policy unchanged and can enforce constraints that were not present during training.

Control barrier functions (CBFs) are widely used to construct such safety filters. 
A CBF expresses a constraint as a condition on the control input, and the filter modifies the tracker’s proposed command only as much as needed to satisfy that condition~\cite{ames2017control}. 
Whole-body formulations can additionally account for contact constraints and task priorities when selecting the correction~\cite{morton2026constrained}.
This modularity is appealing, as the tracker can be treated as a fixed nominal controller, while the filter supplies policy-agnostic corrections, seemingly allowing the two components to be designed independently.

\begin{figure}[!t]
\centering
\includegraphics[width=\columnwidth]{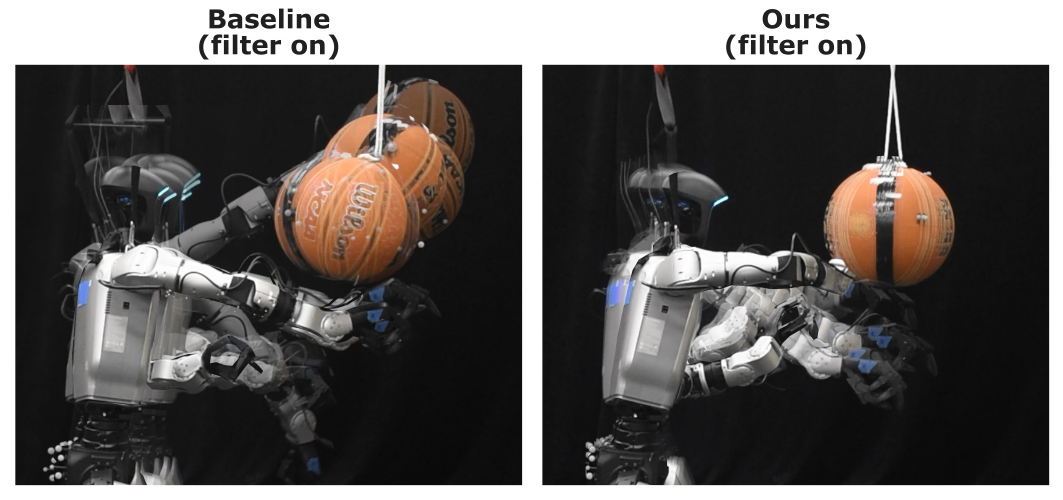}
\caption{Whole-body tracking with an active safety filter (avoid the ball). The continued-training baseline (left) and CoFiT (right) track the same reaching reference. Five of ten baseline trials required an operator stop; CoFiT completed all ten. Overlaid poses are successive stills from one trial.}
\label{fig:hero}
\end{figure}

In this work, we show that treating the tracking policy and safety filter independently induces fundamental mismatches in closed-loop execution (Fig. ~\ref{fig:hero}). 
The safety filter changes the tracker’s outputs even though the tracker was trained under an unfiltered closed loop. 
As a result, both the actions executed by the robot and the states induced by those actions differ from those encountered during training.
Through systematic case studies, we identify three key sources of incompatibility at the policy–filter interface: \textit{dynamics mismatch}, \textit{objective mismatch}, and \textit{information mismatch}. 
Guided by these findings and their underlying causes, we develop \textit{CoFiT} (\underline{Co}nstrained \underline{Fi}lter-aware \underline{T}uning), a method that adapts a pretrained tracker to an existing CBF filter. 
CoFiT fine-tunes the tracker with filters active and supplies it with compact constraint information, temporal feedback, and constraint-aware penalties. 
This allows the tracker to anticipate filter interventions and respond with coordinated, constraint-compatible motion while retaining its original tracking objective where possible.

\smallskip

\noindent Our contributions are as follows:
\begin{itemize}
    \item We provide a systematic analysis of the policy--filter interface, identifying dynamics, objective, and information mismatches between pretrained tracking policies and runtime CBF filters and relating them to specific training and interface design choices.
    \item We introduce \textit{CoFiT}, a filter-aware fine-tuning method that adapts pretrained whole-body tracking policies to an existing CBF filter using action-space constraint observations, temporal feedback, and constraint-aware rewards.
    \item We validate CoFiT through extensive ablations and experiments across multiple base trackers in simulation and on hardware, demonstrating fewer constraint violations, smaller filter corrections, and smoother motion.
\end{itemize}

\section{Related Work}
\label{sec:related-work}
We review prior work along two complementary directions: learning in the presence of runtime safety filters, and incorporating safety constraints across the control hierarchy for legged and humanoid robots.

\subsection{Learning with safety filters}
Safety filters separate nominal task execution from explicit safety enforcement by admitting or modifying a policy's proposed actions according to safety constraints.
CBF-based methods enforce local model-based inequality constraints on proposed controls, while predictive safety filters evaluate candidate inputs through backup trajectories and terminal safe sets~\cite{cheng2019endtoend,wabersich2021a}.
Several works train policies directly on the resulting filtered system so that they can adapt to these interventions.
CBF-RL combines training-time filtering with rewards based on barrier conditions and filter corrections~\cite{yang2026cbfrl}, while Safety Filtering While Training learns from filtered transitions and penalizes corrective interventions with a predictive filter retained during evaluation~\cite{bejarano2025safety}.
Related approaches differentiate through CBF layers~\cite{yang2023differentiable} or expose barrier and Lie-derivative quantities to the policy when selecting CBF-related parameters~\cite{wijayatunga2026learning}.
These methods show that policies can benefit from being trained with knowledge of the safety mechanism, but they do not address how a pretrained whole-body tracker should be adapted to an already deployed runtime filter, nor which constraint representations and feedback signals are most useful for doing so. 
Our work targets this setting by conditioning the nominal whole-body tracker on a compact action-space representation of constraint information and systematically studying how this constraint information, along with temporal feedback and training objectives, affects behavior under the inherited filter.

\subsection{Constraints across the control hierarchy}
Safety constraints can be incorporated at multiple levels of a legged or humanoid control stack, including motion planning, reference generation, and whole-body command execution. 
Planning and tracking layers can share CBF constraints, while filters above learned locomotion controllers can modify high-level references~\cite{grandia2021multilayered,yang2025shield}.
At the command level, humanoid controllers combine barrier constraints with full dynamics and contact variables~\cite{khazoom2022humanoid,paredes2023safe}.
Task-consistent formulations further account for how safety corrections affect the primary motion and posture, as in OSCBF for manipulation and ConstrainedMimic for pretrained humanoid tracking~\cite{morton2026constrained, morton2025safe}.
Other works study the role of safety-related perception. For example, PAC-MAN combines depth with barrier-based rewards for reactive humanoid dodgeball ~\cite{yang2026pacman}.
Collectively, these works establish effective mechanisms for injecting safety constraints at different levels of the control hierarchy, but largely treat the nominal tracking policy and downstream whole-body filter as separately designed components. 
Our work focuses on this interface: we retain the existing whole-body safety filter and study how the tracking policy that supplies its nominal commands should be adapted to account for the filter's constraints and interventions.

\section{Preliminaries}
\label{sec:preliminaries}
In this section, we first briefly review the relative-degree-two CBF formulation used to encode safety constraints (Section~\ref{sec:relative_degree_two_cbf}), and then describe the whole-body filtering framework through which these constraints modify the tracking policy's nominal commands (Section~\ref{sec:whole-body-constrained-tracking}).

\subsection{Relative Degree Two Control Barrier Functions}
\label{sec:relative_degree_two_cbf}
A control barrier function expresses a state constraint as a condition on the control input~\cite{ames2017control}.
Let $h_i(z)\geq0$ define an admissible region for a control-affine model $\dot z=f(z)+g(z)u$, with Lie derivatives $L_fh_i=\nabla h_i^{\mathsf T}f$ and $L_gh_i=\nabla h_i^{\mathsf T}g$ describing drift and input sensitivity.
Position constraints under torque control have $L_gh_i=0$, since the input enters through acceleration.
Such constraints have relative degree two and can be enforced through:
\begin{equation}
\begin{aligned}
h_{2,i}&=L_fh_i+\alpha(h_i),\\
L_fh_{2,i}&+L_gh_{2,i}u+\alpha_2(h_{2,i})\geq0,
\end{aligned}
\label{eq:prelim-hocbf}
\end{equation}
where $\alpha,\alpha_2$ are extended class-$\mathcal K_\infty$ functions~\cite{morton2026constrained, xiao2019control}. \footnote{See \cite{xiao2019control} for a complete construction of higher-order CBFs.}
The intermediate barrier $h_{2,i}$ combines the barrier value with approach rate, allowing the CBF condition to regulate not only proximity to the constraint boundary but also how rapidly the system approaches it.

\subsection{Whole-Body Constrained Tracking}
\label{sec:whole-body-constrained-tracking}
A whole-body filter should choose a constraint correction without unnecessarily disturbing contact support or the motion task.
Operational-space formulations therefore measure intervention according to its effects on the task rather than through a uniform distance in torque space~\cite{morton2025safe}. %
ConstrainedMimic extends this principle to contact-consistent humanoid dynamics by weighting changes in contact forces, task accelerations, and posture accelerations~\cite{morton2026constrained}. %
We use this framework to construct all safety filters in this work.

For a spherical body proxy $i$ and a static spherical obstacle $j$, we define the distance barrier:
\begin{equation}
h_{ij}(q)=\|p_i(q)-c_j\|_2-r_i-r_j-d_{\mathrm{safe}},
\label{eq:prelim-sphere}
\end{equation}
where $p_i(q)$ and $c_j$ denote the sphere centers, $r_i$ and $r_j$ their radii, and $d_{\mathrm{safe}}\geq0$ the desired safety buffer.
For other obstacle shapes, we use corresponding signed-distance functions.

The torque input $u=\tau$ enters the condition through the contact-consistent model.
Each CBF constraint yields a row $G_{i,t}=-L_gh_{2,i}$ with $b_{i,t}=L_fh_{2,i}+\alpha_2(h_{2,i})$.
Stacking these constraints together with actuator bounds gives the quadratic program:
\begin{equation}
\begin{aligned}
\underset{\tau,\,\delta}{\operatorname{minimize}}\quad
&\tfrac12\Delta\tau^{\mathsf T}W_t\Delta\tau
 +\lambda_{\mathrm{QP}}^{\mathsf T}\delta\\
\text{subject to}\quad &G_t\tau\leq b_t+\delta,\\
&\Delta\tau=\tau-\tau_t^{\mathrm{nom}}\\
&\delta\geq0.
\end{aligned}
\label{eq:prelim-filter}
\end{equation}
Here, $\tau_t^{\mathrm{nom}}$ denotes the bounded nominal torque proposed by the tracker, $\lambda_{\mathrm{QP}}>0$ penalizes constraint relaxation, and $W_t\succeq0$ is the task metric that weighs corrective effects on contact force and task/posture accelerations.
The solution $(\tau_t^{\mathrm F},\delta_t^\ast)$ provides the filtered torque and slack variables. The filtered torque is then mapped back to joint targets.

Given this fixed runtime filter, our objective is to fine-tune the pretrained tracker so that its nominal commands lead to better constraint adherence and require smaller corrective interventions, while limiting losses in tracking accuracy and motion quality (smooth, coordinated motion without disruptive oscillations or chattering). Because zero-order hold, model uncertainty, and QP relaxation can still produce violations during execution, we evaluate realized constraint satisfaction empirically rather than claim an unconditional safety guarantee. To respect the planner–tracker abstraction, we focus on handling local interference while maintaining coordinated motion, rather than expecting even a fine-tuned tracker to replan around fully blocking obstacles.

\begin{figure}[t]
\centering
\includegraphics[width=\columnwidth]{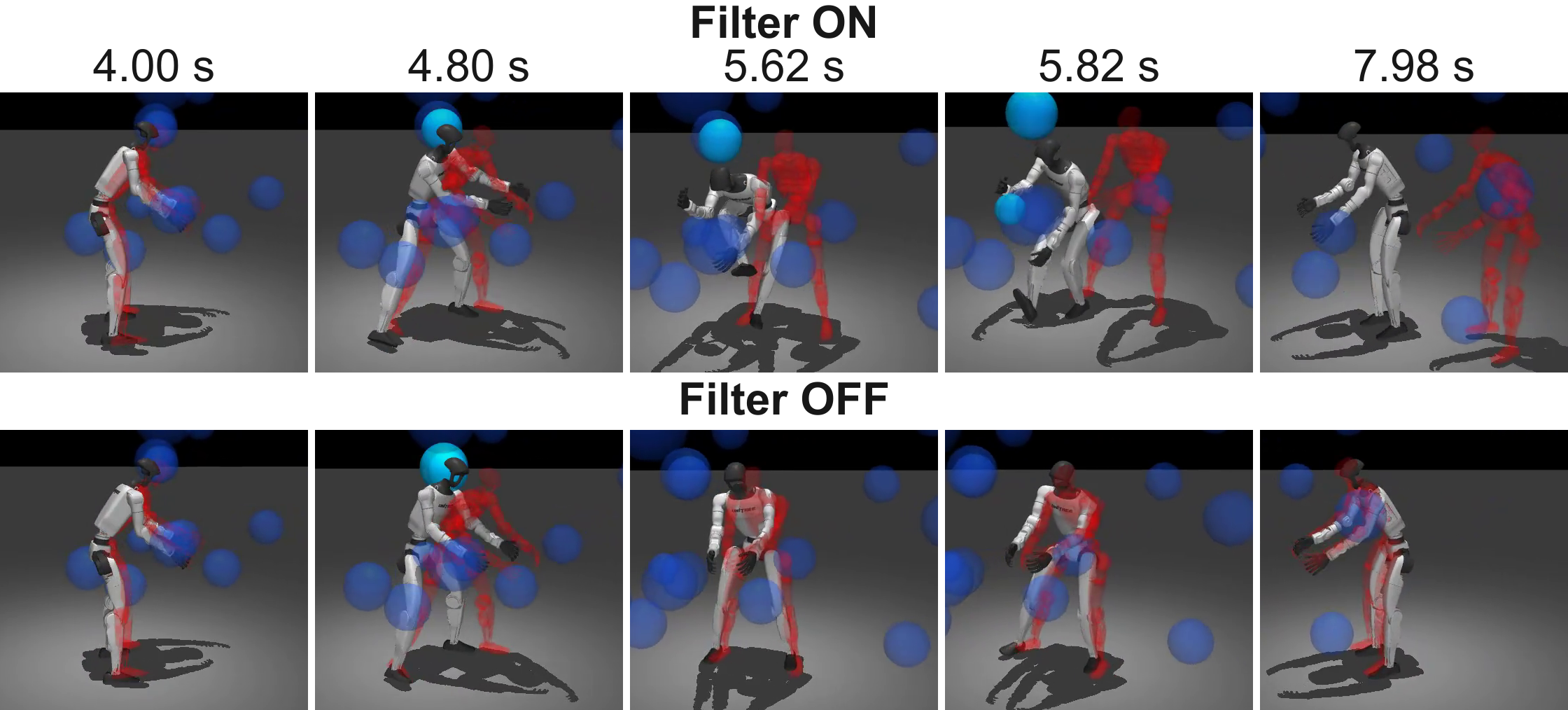}
\caption{Tracking with the filter on (top) and off (bottom) with the same base policy and scene. White: robot. Red: reference. Cyan: barrier violation.}
\label{fig:dynamics-mismatch}
\end{figure}

\section{Failure Modes of Independently Designed Policies and Safety Filters}
\label{sec:failure-modes}
Base trackers were trained with a nominal unconstrained model of the actuators, but applying filters at runtime introduces a distribution shift that impacts the tracking and stability performance of the underlying policy.
We distinguish three interacting mismatches in the composed policy-filter loop, regarding dynamics, objective, and information.

\subsection{Dynamics Mismatch}
\label{sec:dynamics-mismatch}
CBFs produce endogenous and autocorrelated perturbations on unfiltered dynamics.
Prior work has shown that CBF-QPs can prevent convergence to a safe target by introducing stable equilibria elsewhere~\cite{mestres2025control}, or maintain barrier satisfaction while state components not directly constrained by the barrier diverge~\cite{choi2025when}.
For whole-body humanoid tracking, the corresponding concerns are whether enforcing clearance preserves both progress toward the reference and whole-body stabilization.

This can be particularly problematic.
Dynamic balance requires coordinated posture and ground support beyond what a torso-clearance barrier specifies.
The task-space QP addresses this by distributing corrections across joints while limiting changes to contact forces, motion, and posture (Section~\ref{sec:whole-body-constrained-tracking}).
But minimizing this instantaneous intervention does not guarantee that the pretrained tracker will maintain coordination in its subsequent response.
Fig.~\ref{fig:dynamics-mismatch} (4.80--5.62\,s) shows an illustration of why that response matters.
The torso-CBF interaction is followed by a sharp forward bend and raised knee, unlike the unfiltered comparison.
The raised knee also highlights how a disturbed response can affect the filter itself.
Mismatched dynamics can cause frequent and unexpected contact transitions, leading to increased uncertainty over contact mode (and thus which support dynamics model to use), which can further disturb balance~\cite{morton2026constrained}.

With multiple distinct constraints, these issues compound. In Fig.~\ref{fig:dynamics-mismatch} (5.82\,s), the hand encounters another constraint while this whole-body response is still occurring.
The motion used to regain coordination is itself subject to correction, so the tracker must recover under continuing intervention.
By Fig.~\ref{fig:dynamics-mismatch} (7.98\,s), the robot is more upright but remains displaced from its reference.
This instability clearly motivates fine-tuning with the filter active, to help the tracker learn coordinated responses to interventions~\cite{bejarano2025safety}.
More subtly, this supports pairing correction and motion histories to help the tracker account for the compounding effects of filter interventions.

\subsection{Objective Mismatch}
\label{sec:objective-mismatch}
A tracker's reference pursuit can often be incompatible with safety, and while the filter can suppress instantaneously unsafe actions, it cannot guide the policy's objective towards safety.
For example, in the forward reach of Fig.~\ref{fig:hero} (left), the reference enters the obstacle.
Keeping the hand outside yields a tracking error that encourages the policy to reach forward again.
Prior work identifies such persistent controller--filter opposition as a source of chattering and oscillations~\cite{bejarano2025safety}, which we observe in our example.
The baseline continues forward with violent oscillations and displaces the ball. And, even in cases without constraint violations (not pictured), we observe pronounced jerky motion, illustrating how persistent controller--filter opposition can compound the whole-body disturbances discussed in Section~\ref{sec:dynamics-mismatch}.
The task is therefore not simply to prevent violation, but to accommodate the blocked reach without repeatedly disrupting the coordination needed for the rest of the motion.

Training on filtered transitions alone does not explicitly reward the policy for reducing intervention.
Under filter constraints, different proposals can yield the same executed command and execution-based reward~\cite{markgraf2026safe}.
For a filter-blocked trajectory, this means that proposing motion further into the restriction can be indistinguishable in that reward from proposing less, even though more correction is required.
In the relaxed QP~\eqref{eq:prelim-filter}, stronger incompatibility can instead increase constraint relaxation without increasing correction, and so the correction cost alone does not penalize increased slack in the solution.
These distinctions motivate adding penalties to incompatible proposals and costly corrections, following CBF-RL~\cite{yang2026cbfrl}, and additionally penalizing solved slack so the learner is discouraged from leaving barrier conditions relaxed, not merely from demanding large corrections.
This also motivates using the \emph{filtered} previous action in policy feedback and in any policy-native action-rate penalty.

\subsection{Information Mismatch}
\label{sec:information-mismatch}
Simply rewarding compatibility does not tell the tracker which actions the current constraints permit.
Consider a hypothetical variation of Fig.~\ref{fig:hero}: hold the robot state, reference, and complete native input history fixed, but place the sphere just left or right of the upcoming hand path.
Before either placement elicits correction, the tracker produces the same action distribution, although a lateral adjustment that increases clearance in one scene decreases it in the other.
A penalty can favor avoiding intervention in both scenes, but identical inputs give the policy no basis for distinguishing the placements.
Past corrections cannot supply that distinction before either scene has changed a command. Thus, the policies should be aware of constraint information, even before the constraints become active. Further, the provided information should allow the policy to discern which action directions will lead to violation, as well as constraint state and approach rate (for our example, how far the hand is from the ball and how quickly it approaches). Further, to help alleviate the dynamics mismatches discussed in Sec. \ref{sec:dynamics-mismatch}, we should supply the policy a history of this constraint information.

\begin{figure}[t]
\centering
\includegraphics[width=\columnwidth]{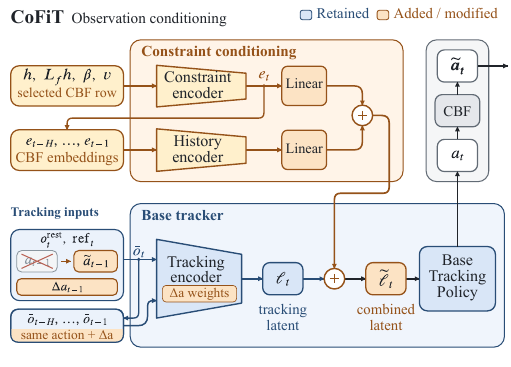}
\caption{CoFiT observation conditioning}
\label{fig:method-overview}
\end{figure}

\section{Filter-Aware Policy Design}
\label{sec:filter-aware-design}
CoFiT (illustrated in Fig.~\ref{fig:method-overview}) fine-tunes whole-body trackers to cooperate with a safety filter while maintaining performance.
We retain the base tracker's MLP structure and native tracking rewards as a backbone, with the ConstrainedMimic whole-body filter active throughout training and deployment~\cite{morton2026constrained}.

\subsection{Constraint-Context Observations}
As motivated in Sec.~\ref{sec:information-mismatch}, we endow the base policy with information regarding which directions would lead to violation, and how urgent the constraint is. 

To give the tracker this relationship directly, we express the inequality constraint in policy-action coordinates using the torque map $\tau=B_ta+c_t$ before target and torque clipping.
Here, $a\in\mathbb R^m$ is the policy action, and $B_t,c_t$ are the gain and offset of the proportional--derivative controller at the current state $z_t$.
Substituting this map into the CBF rows of~\eqref{eq:prelim-filter} gives:

\begin{equation}
\begin{aligned}
v_{i,t}^{\mathsf T}a+\beta_{i,t}&\geq0,\\
v_{i,t}^{\mathsf T}&=-G_{i,t}B_t,\\
\beta_{i,t}&=b_{i,t}-G_{i,t}c_t.
\end{aligned}
\label{eq:method-affine-condition}
\end{equation}
The constraint normal $v_{i,t}$ points toward action changes that push the system away from the constraint while $\beta_{i,t}$ is the constraint margin at zero action.

We also include the barrier value $h_i$ and rate $L_fh_i$, which describe the constraint's current state and approach.

The resulting observation is
\begin{equation}
\begin{aligned}
\chi_{i,t}&=\left[h_i; \ L_fh_i;\
\widehat\beta_{i,t};\ \widehat v_{i,t}\right]
\in\mathbb R^{m+3},
\end{aligned}
\label{eq:method-context-row}
\end{equation}
where hats denote divisions by $\|v_{i,t}\|_2$ to normalize inputs.

This requires no additional expensive differentiations, since the QP already computes this information.
CoFiT selects the constraint with the smallest barrier value and encodes it into $e_t$ with a small MLP (the filter continues to enforce all constraints). Because this observation does not rely on specific constraint geometry, it shares meaning across \emph{any relative degree two CBF}.

\subsection{Temporal Feedback and Policy Integration}
Constraint history describes how local restrictions evolve along the motion, while correction history records how the filter changed the tracker's proposals.
Let $q_t^{\mathrm{prop}}$ denote the target after action scaling, offset, and target clipping, but before correction.
Comparing it with the returned $q_t^{\mathrm{cmd}}$ gives the feedback correction and action:
\begin{equation}
\Delta a_t=\frac{q_t^{\mathrm{cmd}}-q_t^{\mathrm{prop}}}{s_a},
\qquad\widetilde a_t=a_t+\Delta a_t,
\label{eq:method-action-delta}
\end{equation}
where division by the action scale $s_a$ is elementwise.
We use $\widetilde a_{t-1}$ in the native previous-action channel and as the reference for the action-rate term $a_t-\widetilde a_{t-1}$, so both refer to command feedback after filtering.

We append the latest correction $\Delta a_{t-1}$ to the native observations to form $\bar o_t$, and retain the $H=10$ preceding frames of both $\bar o_t$ and the encoded context $e_t$.
Both histories reset at episode boundaries, and stored context embeddings retain their collection-time values without gradient flow. 

The current context and its encoded history are added to the first hidden layer of the base policy (See Fig.~\ref{fig:method-overview}).
We initialize all new weights to zero so that before any fine-tuning, the augmented network behaves identically to the base policy.
All of these changes apply to both the actor and the critic (each learning a separate network).

\subsection{Policy--Filter Alignment Rewards}
We add three filter-related penalties to the base policy. The first two are adaptations of CBF-RL's correction and raw-condition rewards, modified for a whole-body task-space filter metric~\cite{yang2026cbfrl}. Thirdly, because the QP in~\eqref{eq:prelim-filter} may relax constraints when the problem would otherwise be infeasible, we penalize the largest solved slack.

Letting $\Delta\tau_t=\tau_t^{\mathrm F}-\tau_t^{\mathrm{nom}}$ be the torque-space filter intervention and using the QP rows from (\ref{eq:prelim-filter}), the added penalties are:
\begin{equation}
\begin{aligned}
r_t^c&=\exp(\frac{-\Delta\tau_t^{\mathsf T}W_t\Delta\tau_t}{\sigma^2})-1,\\
r_t^r&=\min ~\left\{ 0,~\min_{i\in\mathcal V_t} \left(b_{i,t}-G_{i,t}\tau_t^{\mathrm{nom}} \right)  \right\} ,\\
r_t^s&=-\max_{i\in\mathcal V_t}~(\delta^*_{i,t}).
\end{aligned}
\label{eq:method-alignment-terms}
\end{equation}
Here, $\sigma>0$ sets the correction scale, $\mathcal V_t$ contains CBF constraint rows, and $\delta^*_{i,t}$ is the solved slack.

Retaining native reward terms and adding these penalties, our reward becomes:
\begin{equation}
\bar r_t=r_t^{\mathrm{native}}+\lambda_c r_t^c+\lambda_r r_t^r+\lambda_s r_t^s.
\label{eq:method-combined-reward}
\end{equation}

\newsavebox{\spherefigure}\newsavebox{\spheretable}
\begin{lrbox}{\spherefigure}\begin{minipage}{\columnwidth}
\includegraphics[width=\linewidth]{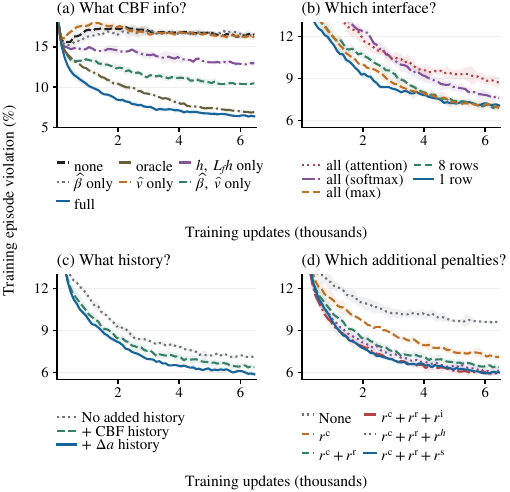}
\captionof{figure}{Curves average three seeds after 100-update windows, with bands showing sample SD.}
\label{fig:sphere-design}
\end{minipage}\end{lrbox}
\begin{lrbox}{\spheretable}\begin{minipage}{\columnwidth}
\captionof{table}{Validation on sphere scenes with the runtime filter. Means with sample SD in subscripts. Lower is better.}
\label{tab:sphere-recipe}
\fontsize{8}{9.6}\selectfont
\setlength{\tabcolsep}{1.5pt}
\renewcommand{\arraystretch}{1.12}
\begin{tabular*}{\linewidth}{@{\extracolsep{\fill}}llrrr@{}}\toprule
& & Violation & Correction & World MPJPE\\
& Condition & time (\%) & RMS (rad) & Filtered (cm)\\\midrule
\multirow{3}{*}{\rotatebox[origin=c]{90}{\scriptsize\itshape Baselines}}
& Pretrained & 4.13\textsubscript{\scriptsize$n{=}1$} & 0.321\textsubscript{\scriptsize$n{=}1$} & 5.47\textsubscript{\scriptsize$n{=}1$}\\
& Unfiltered training & 5.25\sd{0.51} & 0.372\sd{0.016} & 5.30\sd{0.16}\\
& Filter-only & 8.00\sd{0.24} & 0.540\sd{0.012} & 6.02\sd{0.03}\\
\midrule
\multirow{3}{*}{\rotatebox[origin=c]{90}{\scriptsize\itshape Interface}}
& 1 constraint & 1.22\sd{0.03} & 0.191\sd{0.010} & 5.88\sd{0.04}\\
& 8 constraints & 1.63\sd{0.07} & 0.185\sd{0.010} & 5.95\sd{0.07}\\
& Pooled & 1.55\sd{0.23} & 0.178\sd{0.010} & 6.07\sd{0.03}\\
\midrule
\multirow{3}{*}{\rotatebox[origin=c]{90}{\scriptsize\shortstack{\textit{Penalties}\\Added to\\$r^{\mathrm c}+r^{\mathrm r}$}}}
& + intermediate & 0.88\sd{0.05} & 0.201\sd{0.004} & 6.36\sd{0.10}\\
& + violation & 0.97\sd{0.06} & 0.204\sd{0.000} & 6.37\sd{0.08}\\
& + slack & 0.94\sd{0.04} & 0.197\sd{0.004} & 6.37\sd{0.12}\\
\midrule
& \textbf{CoFiT} & 0.58\sd{0.02} & 0.173\sd{0.010} & 6.25\sd{0.06}\\
\bottomrule\end{tabular*}

\par\vspace{4pt}{\footnotesize Three training seeds except pretrained ($n=1$). Each seed evaluates 4,096 initial episodes through first termination. Violation time is the percent of recorded steps with any violation; correction RMS is the joint-command correction norm.\par}
\end{minipage}\end{lrbox}
\newcommand{\spherepair}{\noindent\begin{minipage}[t]{\columnwidth}\vspace{0pt}\usebox{\spherefigure}\end{minipage}\hfill\begin{minipage}[t]{\columnwidth}\vspace{0pt}\usebox{\spheretable}\end{minipage}}
\typeout{COLUMN-WIDTH=\the\columnwidth}
\typeout{FIGURE-HEIGHT=\the\dimexpr\ht\spherefigure+\dp\spherefigure\relax}
\typeout{TABLE-HEIGHT=\the\dimexpr\ht\spheretable+\dp\spheretable\relax}

\begin{figure}[!t]
  \centering
  \usebox{\spherefigure}
\end{figure}

\begin{table}[!t]
  \centering
  \usebox{\spheretable}
\end{table}

\section{Experiments}
\label{sec:experiments}
\providecommand{\CoFiTSphereViolationDrop}{89}
\providecommand{\CoFiTSphereDepthDrop}{5.2}
\providecommand{\CoFiTSphereTrackingCost}{0.9}
\providecommand{\CoFiTHardwareViolationDrop}{83}
\providecommand{\TODO}[1]{\textbf{[AUTHOR INPUT: #1]}}
We evaluate CoFiT through four outcomes: realized safety, cooperation with the filter, motion quality, and reference tracking. We first isolate design choices in a simplified setting, then validate on more realistic scenes, a second tracker, and hardware.

We provide the tracker with fixed world-frame trajectories that cannot replan in response to tracking errors or safety interventions. Our scenes are therefore designed as near-misses or mild interferences requiring local readjustment rather than rerouting. Because drift in lateral position or heading may turn these into planning-level problems, we use a world-position-aware tracker trained from scratch with TWIST2's recipe. We observed less native drift with SONIC and, given its higher pretraining cost, use its publicly available tracker module to initialize fine-tuning~\cite{ze2025twist2,luo2026sonic}.

TWIST2 training uses a custom MJLab environment~\cite{zakka2026mjlab}, while SONIC is fine-tuned in its native Isaac Lab~\cite{mittal2025isaac}. Both use filters constructed in CBFpy~\cite{morton2025safe} and an elastic solver adapted from QPAX~\cite{tracy2024on}.
\paragraph{Metrics} We measure safety by violation time, the percentage of recorded samples with any negative physical clearance; filter cooperation by correction RMS, the root-mean-square \(L_2\) norm of joint-target corrections. Motion quality is measured by residual acceleration as the RMS acceleration of joint-position tracking error, computed by second differences, and by command total variation (TV), changes between consecutive applied joint targets. Tracking accuracy is measured by world-frame mean per-joint position error (MPJPE) over the tracked key-body positions.
\subsection{Controlled Design Study on Sphere Scenes}
\label{sec:sphere-ablation}
We sample motions from TWIST2's training corpus and place spherical obstacles along their reference trajectories, close enough that the tracker must accommodate them while continuing the motion
(see example in Fig.~\ref{fig:dynamics-mismatch}).
We compare 29 configurations over three training seeds each to specify the details of CoFiT's observation, history, and reward design (following the broad insights from Section~\ref{sec:failure-modes}), and report the most informative comparisons.
While we care about more than safety, it remains the first priority. Thus, we use the training violation-rate curves in Fig.~\ref{fig:sphere-design} as our primary selection tool, deferring to other metrics only as needed. Note that training violations are calculated under exploration noise and domain randomization, so their values are higher than and not directly comparable to deterministic validation numbers in Table~\ref{tab:sphere-recipe}.

\paragraph{What constraint information does the policy need? (Fig.~\ref{fig:sphere-design}a)} We ablate subsets of the observation package given in~\eqref{eq:method-context-row}, with the encoder, constraint history, and rewards fixed. We find neither half of the affine constraint to be useful alone: supplying only the condition margin $\hat\beta$ or only the action-space constraint normal $\hat v$ performs no better than supplying no constraint information at all. Supplied together, they substantially reduce training violations. The barrier value and rate also help on their own: they describe the current clearance and approach, but do not specify how candidate actions affect the constraint. The complete package achieves the lowest training violation among the tested subsets. It also reaches lower violation, and does so faster, than the oracle-observation baseline given complete world-frame knowledge of the obstacles. These results support the joint use of constraint-state and action-condition information in this setting.

\paragraph{How should constraint information be integrated? (Fig.~\ref{fig:sphere-design}b)}
We compare ways of integrating the constraint observation with rewards fixed and no constraint history. Attention queries all the encoded constraints using the remaining policy input. Softmax and max pooling aggregate all encoded constraints. Top-$k$ selects the $k$ constraints with the smallest barrier values and encodes them directly, zero-padding when fewer valid constraints are available. Attention and softmax yield the highest training violation rates. The remaining three are essentially tied on training safety. We choose top-$1$ as it is slightly safer in validation (Table~\ref{tab:sphere-recipe}) and simpler. 

\paragraph{What historical information is useful? (Fig.~\ref{fig:sphere-design}c)}
Both base trackers are provided a short window of proprioceptive history, which lets a policy infer unobserved properties of its own dynamics from how the body responded to recent commands~\cite{zhang2025track,kumar2021rma}. As discussed in Sec.~\ref{sec:dynamics-mismatch}, the filter perturbs these dynamics. Holding the nearest-constraint description and rewards fixed, we cumulatively add constraint history, then correction history (the tracker already receives the filtered previous action and a single step of the correction). We find that each addition reduces violation time. 

\paragraph{How should the reward be modified? (Fig.~\ref{fig:sphere-design}d)} 
We ablate the added penalties while retaining the complete nearest-constraint description and its history. Penalizing the task-space correction the filter applies ($r^c$) captures most of the available gain. Adding a penalty on proposals that violate the condition before correction ($r^r$) helps further (nontrivially). Beyond those two, the benefits are incremental: three further penalties, on the intermediate condition ($r^i$), the realized violation ($r^h$), and the largest slack the solver allowed ($r^s$), land close to one another and close to the pair (Table~\ref{tab:sphere-recipe}), with $r^i$ marginally safest (each new reward was weighted to contribute similarly). What matters is that the policy receive \emph{some} signal about what the filter had to do. We choose to use $r^s$, since it alone measures the solver's behavior rather than the robot's state. A practitioner may reasonably choose otherwise. 

\paragraph{The CoFiT recipe}
Thus, these design ablations shape our chosen recipe for CoFiT: the nearest constraint's complete description, histories of that description and of the filter's corrections, and the correction, raw-condition, and slack penalties. Against continued unfiltered training it reduces violation time by
\CoFiTSphereViolationDrop\% and the correction the filter must apply by more than half, for
\CoFiTSphereTrackingCost\,cm more world-frame tracking error
(Table~\ref{tab:sphere-recipe}).

\begin{table*}[t]
\centering
\caption{Constrained tracking in generated scenes. Mean with sample SD in subscripts across training seeds.}
\label{tab:generated-scenes}
\footnotesize
\setlength{\tabcolsep}{2pt}
\renewcommand{\arraystretch}{1.0}
\begin{tabular*}{\textwidth}{@{}ll@{\extracolsep{\fill}}rrrrrr@{}}
\toprule
Backbone & Method & Viol. & Max depth & Correction & World MPJPE & TV & Resid. acc.\\
 & & (\%) & (p95 cm) & (rad RMS) & (cm) & (rad/step) & (rad/s$^2$ RMS)\\
\midrule
TWIST2 & CoFiT & \textbf{1.71\sd{0.10}} & \textbf{8.27\sd{0.55}} & \textbf{0.188\sd{0.003}} & 4.33\sd{0.06} & \textbf{0.174\sd{0.003}} & \textit{18.8\sd{0.3}}\\
 & Observations-only & \textit{5.01\sd{0.22}} & 13.50\sd{0.27} & \textit{0.251\sd{0.007}} & \textbf{3.54\sd{0.04}} & 0.197\sd{0.007} & \textbf{18.3\sd{0.5}}\\
 & Penalties-only & 15.48\sd{0.94} & \textit{11.34\sd{0.64}} & 0.343\sd{0.013} & 5.45\sd{0.24} & \textit{0.180\sd{0.014}} & 19.3\sd{0.4}\\
 & Filter-only & 19.31\sd{0.58} & 14.77\sd{0.28} & 0.458\sd{0.016} & \textit{3.77\sd{0.08}} & 0.259\sd{0.014} & 19.6\sd{0.1}\\
 & Unfiltered training & 23.75\sd{0.27} & 15.58\sd{1.05} & 0.857\sd{0.023} & 4.08\sd{0.03} & 0.212\sd{0.007} & 24.4\sd{0.8}\\
\addlinespace[2pt]
SONIC & CoFiT & \textbf{11.27\sd{0.35}} & 5.31\sd{0.40} & \textbf{0.369\sd{0.021}} & 13.07\sd{0.28} & \textbf{0.158\sd{0.004}} & \textbf{15.5\sd{0.5}}\\
 & Observations-only & \textit{12.40\sd{0.52}} & \textbf{4.95\sd{0.11}} & \textit{0.405\sd{0.013}} & \textit{12.53\sd{1.30}} & \textit{0.162\sd{0.001}} & \textit{15.6\sd{0.1}}\\
 & Penalties-only & 13.38\sd{0.38} & \textit{5.20\sd{0.19}} & 0.495\sd{0.003} & 13.57\sd{0.38} & 0.196\sd{0.005} & 16.9\sd{0.3}\\
 & Filter-only & 14.20\sd{0.90} & 5.69\sd{0.81} & 0.517\sd{0.016} & 12.63\sd{0.66} & 0.200\sd{0.002} & 16.3\sd{0.7}\\
 & Unfiltered training & 20.45\sd{0.28} & 9.25\sd{0.49} & 1.202\sd{0.008} & \textbf{7.11\sd{0.24}} & 0.457\sd{0.008} & 32.9\sd{0.6}\\
\bottomrule
\end{tabular*}
\par\vspace{2pt}
\begin{minipage}{\textwidth}
\scriptsize
Three training seeds per method, 2,048 validation episodes per seed.
\textbf{Bold}/\textit{italic} denote lowest/second-lowest means within each backbone.
SONIC/TWIST2: 2,500/6,500 updates.
\end{minipage}
\end{table*}

\begin{figure}[t]
\centering
\includegraphics[width=\columnwidth]{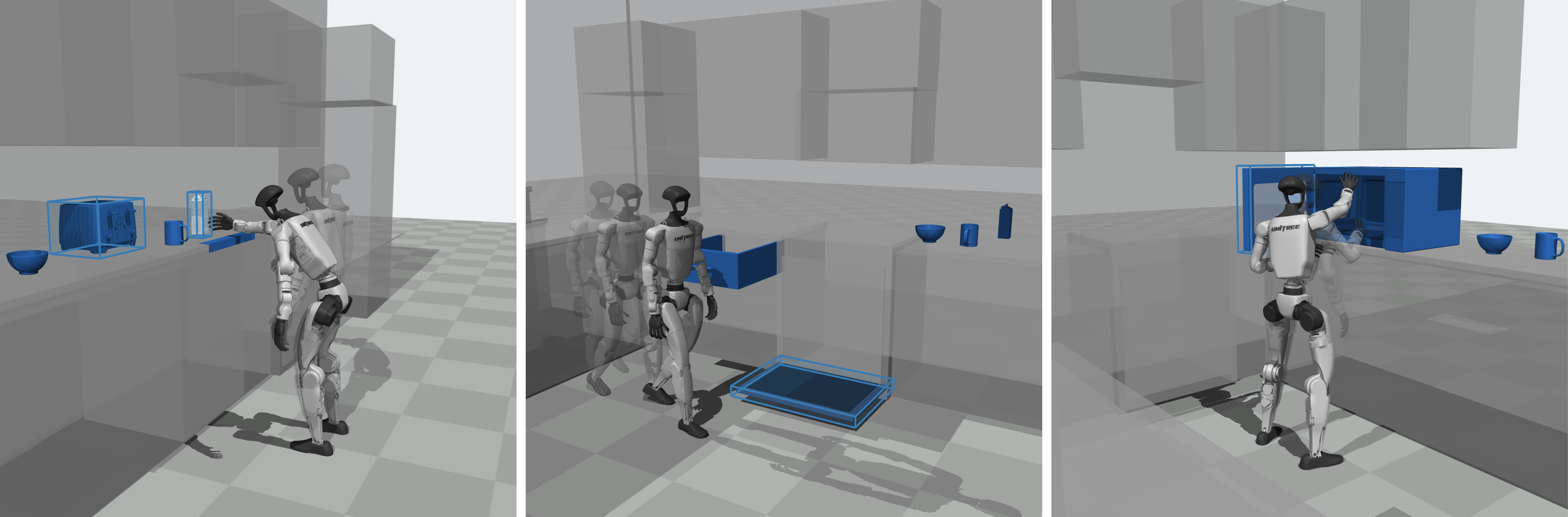}
\caption{Generated constraint scenes with overlaid reference poses and task-relevant obstacles highlighted in blue.}
\label{fig:generated-scenes}
\end{figure}

\begin{figure*}[!t]
  \centering
  \includegraphics[width=\linewidth]{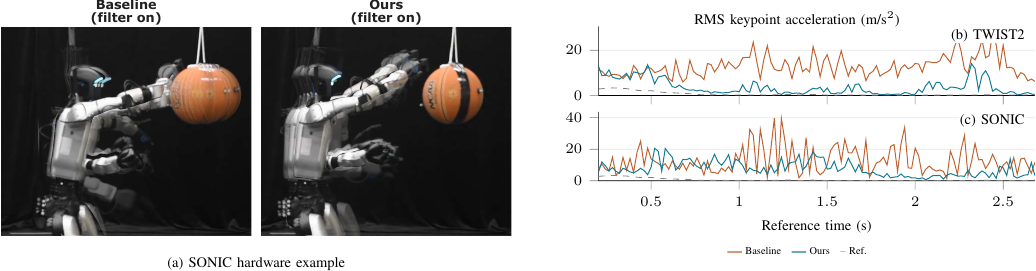}
  \caption{Hardware examples with the filter active.
  (a) SONIC overlay comparison.
  The TWIST2 overlay appears in Fig.~\ref{fig:hero}.
  (b,c) RMS keypoint acceleration summarizes acceleration of tracked body points, measured during the pictured trials.}
  \label{fig:hardware-motion}
\end{figure*}

\begin{table*}[!t]
  \centering
  \caption{Repeated hardware trials.
  Mean with sample SD in subscripts.}
  \label{tab:hardware-reach}
  
\newcommand{\compactrows}{%
TWIST2 & \textbf{CoFiT} & $8.13\sd{1.98}$ & $0.449\sd{0.055}$ & $7.55\sd{1.04}$ & $0.381\sd{0.057}$ & $31.9\sd{2.3}$ & $2.16\sd{0.30}$ & 0/10\\
 & Unfiltered Training & $46.85\sd{30.64}$ & $1.362\sd{0.649}$ & $9.38\sd{3.18}$ & $0.894\sd{0.445}$ & $65.6\sd{27.5}$ & $3.29\sd{1.11}$ & 5/10\\
\addlinespace[2pt]
SONIC & \textbf{CoFiT} & $5.24\sd{2.76}$ & $0.750\sd{0.095}$ & $12.17\sd{3.22}$ & $0.551\sd{0.083}$ & $18.3\sd{3.7}$ & $1.05\sd{0.21}$ & 0/10\\
 & Unfiltered Training & $7.64\sd{3.37}$ & $0.801\sd{0.109}$ & $5.45\sd{0.92}$ & $0.534\sd{0.114}$ & $19.8\sd{4.1}$ & $1.34\sd{0.34}$ & 0/10\\
}

\begingroup
\centering\fontsize{8}{9}\selectfont
\setlength{\tabcolsep}{2pt}\renewcommand{\arraystretch}{.95}
\begin{tabular*}{\linewidth}{@{\extracolsep{\fill}}llccccccc@{}}
\toprule
& & Violation & Correction & World MPJPE & Command TV & Residual accel. & Arm vel. error$^*$ & E-stops\\
Backbone & Method & (\%) & (rad RMS) & (cm) & (rad/step) & (rad/s$^2$ RMS) & (rad/s RMS) & required$^\dagger$\\
\midrule
\compactrows
\bottomrule
\end{tabular*}
\par\vspace{2pt}
\begin{minipage}{\linewidth}
\fontsize{7}{8}\selectfont
Continuous metrics use $n=9$ for SONIC-CoFiT (one logging failure) and $n=10$ otherwise.
E-stop counts use all ten trials per condition.
The filter is always active.
$^*$Calculated over all joints on the reaching arm.
$^\dagger$Operators used e-stops when the robot thrashed violently.
\end{minipage}
\par
\endgroup

\end{table*}

\subsection{Constrained Tracking Across Scenes and Policies}
\label{sec:generated-scenes}
We next evaluate the CoFiT design on generated constraint scenes by fine-tuning TWIST2 and SONIC separately. 

We use a mix of manipulation, locomotion, and locomanipulation reference motions generated via Kimodo~\cite{rempe2026kimodo}, conditioned on language descriptions of household tasks. To generate scene geometry, we sweep the robot's collision bodies along each reference and place obstacles relative to the resulting volumes, so every scene is built around the motion it constrains. Obstacles are spheres, boxes, and capsules, each paired with the bodies it restricts, drawn from free-space layouts and kitchen geometry from RoboCasa ~\cite{nasiriany2024robocasa} (see Fig.~\ref{fig:generated-scenes}). Unlike the sphere scenes, we let obstacles overlap the reference itself, so the tracker must depart from its prescribed reference to remain safe. We fine-tune each backbone separately over three seeds, comparing CoFiT with observations-only (our observation package with native rewards), penalties-only (our added penalties without constraint observations), filter-only fine-tuning, and continued training without filtering.\footnote{For all remaining TWIST2 results, CoFiT uses $\lambda_c=1$, $\lambda_r=0.005$, $\lambda_s=0.015$, and $\sigma=0.5$. For SONIC we scale the penalty weights by the ratio of average returns between the two trackers.}

We find that CoFiT improves constraint adherence and policy-filter cooperation in both trackers.
With the filter active for all five conditions, CoFiT achieves the lowest mean violation time and correction RMS in both trackers. Relative to filter-only fine-tuning, violation time falls by 91\% for TWIST2 and 21\% for SONIC, and correction RMS by 59\% and 29\%, respectively (Table~\ref{tab:generated-scenes}).
The five treatments also have the same ordering on both measures in the two trackers (CoFiT, observations-only, penalties-only, filtered training, and unfiltered training). %
Further, CoFiT provides smooth motion, giving the lowest mean command variation in both trackers, together with lower residual acceleration than filter-only training.

The observation-only comparison explains much of this improvement.
In both trackers, observations-only outperforms penalties-only on violation and correction. More importantly, adding our observations to filter-only training improves constraint adherence without higher mean world-frame tracking error, while also reducing residual acceleration and command variation, with the native rewards unchanged.
The observations-only tracker therefore learns to demand less correction without an explicit penalty for doing so.

Compared with observations-only training, adding our penalties further reduces violation time, correction magnitude, and command variation in both trackers, at the cost of tracking accuracy. For SONIC, much of the tracking difference from continued unfiltered training is already present after filter-only fine-tuning: world-frame MPJPE is 7.11\,cm after continued unfiltered training, 12.63\,cm after filter-only fine-tuning, and 13.07\,cm with CoFiT. \footnote{We hypothesize, but do not validate experimentally, that SONIC’s lack of world-frame feedback contributes to this loss.}

Across these varied scenes and both trackers, we replicate the main design lesson of the sphere study: constraint observations improve adherence under native rewards, while added penalties further reduce violation time and corrective effort at a tracking cost relative to observations-only training.

\subsection{Real-Robot Validation}
\label{sec:hardware-validation}
Finally, we test whether the learned policy-filter interaction improves execution on a physical Unitree G1 during a prescribed reaching task near a basketball.
For each backbone, we evaluate  CoFiT and the unfiltered-training baseline---both trained on the generated scenes as in Sec.~\ref{sec:generated-scenes}---as they execute the same motion with the whole-body filter active at 50\,Hz. 
The reference enters the modeled exclusion sphere, requiring the tracker to accommodate the blocked reach while maintaining the rest of the whole-body motion. We include a conservative  safety buffer of approximately 4\,cm to allow for sim-to-real mismatch, giving a 32\,cm-diameter virtual obstacle. \emph{This buffer was fixed before the experiments.}
For both trackers, the filter receives the pelvis and basketball poses from motion capture (MoCap), together with robot state for forward kinematics.  

As before, our TWIST2-based policy receives the additional global-position feedback, while SONIC does not.
Table~\ref{tab:hardware-reach} reports ten physical repetitions per condition. One SONIC-CoFiT trial had unusable logged data but completed without an operator stop. It is excluded from continuous-metric statistics but included in the stop count.
With our TWIST2-based tracker on the G1, we demonstrate improved constraint adherence and tracking together with markedly lower residual acceleration and command variation. %
CoFiT reduces violation time by 83\% relative to continued training, while lowering world-frame tracking error and the correction required from the filter (Table~\ref{tab:hardware-reach}).
Across repeated trials, residual acceleration falls from 65.6 to 31.9\,rad/s$^2$, while command variation also falls by more than half.
The RMS keypoint-acceleration trace in Fig.~\ref{fig:hardware-motion}(b) shows much smoother motion for CoFiT than the baseline through the illustrated reach, and the pose overlays in Fig.~\ref{fig:hero} show the resulting difference in whole-body motion.
Five of ten baseline TWIST2 trials required an e-stop due to violent thrashing whereas CoFiT completed all ten without a stop.

Using SONIC on the same G1, we also find reductions in correction RMS, residual acceleration, and reaching-arm velocity error with CoFiT. Mean violation time falls from 7.64\% to 5.24\%. These gains accompany an increase in world-frame MPJPE from 5.45 to 12.17\,cm and slightly higher command variation. \footnote{As hypothesized in Sec.~\ref{sec:generated-scenes}, the lack of direct world-frame feedback may contribute to this tracking loss.}

Both policies remain relatively steady in Fig.~\ref{fig:hardware-motion}(a), although the baseline makes light contact with the ball. The corresponding acceleration traces in Fig.~\ref{fig:hardware-motion}(c) show an improvement in smoothness with CoFiT, though less pronounced than for TWIST2. All ten CoFiT trials complete without an operator stop, as do all ten continued-training trials.

Together, these hardware results show that the gains in constraint adherence and filter cooperation extend to physical execution on both trackers, with every CoFiT trial completing without an operator stop.

\section{ Conclusion} 
\label{sec:discussion}
CoFiT fine-tunes humanoid trackers to improve constraint adherence and cooperation with runtime filters. We find that a compact constraint description already computed by the filter improves cooperation even under native rewards, with correction feedback and penalties providing further gains. The recipe applies across both tested trackers and several obstacle geometries. Future work will integrate CoFiT with safety-aware planners to help humanoids carry out more complex tasks safely.
\section*{Acknowledgments}
This work is supported in part by Generative Bionics, and by the National Artificial Intelligence Research Resource (NAIRR) Pilot and the Anvil supercomputer supported by the National Science Foundation (award NSF-OAC 2005632) at Purdue University.  Pranit Mohnot was supported by an NSF Graduate Research Fellowship. Thanks to Daniel Morton for insightful discussions throughout the project.

\bibliographystyle{IEEEtran}
\bibliography{bibliography/references,bibliography/references-additions}
\end{document}